%% file: main.tex
\documentclass{article} 
\usepackage{arxiv}
\renewcommand{\headeright}{}
\renewcommand{\undertitle}{}
\usepackage[round]{natbib}

\input{math_commands.tex}

\usepackage{hyperref}

\defcitealias{fathomnet-2025}{FathomNet 2025}

\usepackage{amsmath,amssymb,amsthm}
\usepackage{algorithm}
\usepackage{algpseudocode}
\usepackage{esdiff}
\usepackage[utf8]{inputenc} 
\usepackage[T1]{fontenc}    
\usepackage{graphicx}
\usepackage{booktabs}       
\usepackage{amsfonts}       
\usepackage{nicefrac}       
\usepackage{makecell}
\usepackage{mathtools}
\usepackage{microtype}      
\usepackage{multirow}
\usepackage{xcolor}         
\usepackage{subcaption}
\usepackage{tcolorbox}

\definecolor{linkblue}{rgb}{0,0.08,0.45}
\hypersetup{colorlinks=true, linkcolor=linkblue, citecolor=linkblue, urlcolor=linkblue}
\hypersetup{pdftitle={GHiML: VLM-Guided Hierarchical Multi-Label Learning with Applications to Benthic Imagery},
            pdfauthor={Isaac Xu, Devi Ayyagari, Martin Gillis, Benjamin Misiuk, Craig J. Brown, Thomas Trappenberg}}

\newtheorem{theorem}{Theorem}

\definecolor{siglow}{HTML}{238B45}   
\definecolor{sighigh}{HTML}{00441B}  

\newcommand{\sigA}[1]{\textcolor{siglow}{#1}}
\newcommand{\sigB}[1]{\textcolor{sighigh}{\boldmath\bfseries#1}}

\newcommand{\dinv}{\ensuremath{d_{\text{inv}}}}
\newcommand{\fCM}{f_\text{CM}}

\title{GHiML: VLM-Guided Hierarchical Multi-Label Learning with Applications to Benthic Imagery}

\renewcommand{\shorttitle}{GHiML: VLM-Guided Hierarchical Multi-Label Learning}
\author{%
  Isaac Xu\textsuperscript{1}\quad
  Devi Ayyagari\textsuperscript{1}\quad
  H. Martin Gillis\textsuperscript{1}\quad
  Benjamin Misiuk\textsuperscript{1,2}\quad
  Craig J.~Brown\textsuperscript{3}\quad
  Thomas Trappenberg\textsuperscript{1}\\[1.5ex]
  \normalfont
  \textsuperscript{1}Faculty of Computer Science, Dalhousie University, Halifax, NS, Canada\\
  \textsuperscript{2}Departments of Geography and Earth Sciences, Memorial University of Newfoundland, St.\ John's, NL, Canada\\
  \textsuperscript{3}Department of Oceanography, Dalhousie University, Halifax, NS, Canada\\[1ex]
  \texttt{isaac.xu@dal.ca}, \texttt{tt@cs.dal.ca}
}

\date{}

\begin{document}

\maketitle

\begin{abstract}
Vision language models (VLMs) encode substantial expert knowledge within their weights, yet their direct application to specialised classification domains such as benthic imagery remains limited. Meanwhile, vision transformers trained on these tasks rely predominantly on low-level visual latents (colours, textures, and shapes) that lack the higher-level reasoning humans employ for identification. We propose Guided Hierarchical Multi-Label (GHiML) learning, a pipeline that extracts conceptual latent variables from a VLM conditioned on training data, generates sample-relevant captions grounded in those variables, and fuses the resulting textual embeddings with image embeddings via cross-attention and a per-node gated router. This approach provides semantic information complementary to low-level visual features, without requiring the VLM to be fine-tuned or a generative model to synthesise training images. We also introduce the inverted HML score (\(\dinv\)), a similarity score that evaluates hierarchical multi-label predictions through graph traversal distance and a cardinality penalty, providing an interpretable alternative to existing measures, which are insensitive to realistic degradations in sparse, deep hierarchies. Experiments on BenthicNet, FathomNet and iNaturalist show that the router improves macro \(F_1\) over the vision-only baseline by up to 6.7\,pp and \(\dinv\) by up to 21.1\,pp on in-domain benthic data, with gains persisting under distribution shift. Ablations locate the contribution of the text itself: on the multi-label biotic hierarchies, removing the per-sample text signal degrades \(\dinv\) while leaving macro \(F_1\) unchanged, consistent with text acting through prediction cardinality rather than node discrimination, whereas iNaturalist, whose annotations are single-label, loses 18.7\% of macro \(F_1\). Our results support the hypothesis that VLMs contain exploitable specialised knowledge that, when appropriately elicited, can augment classification in domains where direct visual learning is insufficient\footnote{Our code is available at \url{https://github.com/DalhousieAI/llm-augmented-benthic-hml-classification}.}.
\end{abstract}

\section{Introduction}
\label{sec:intro}
For general tasks, vision language models (VLMs) have seen widespread adoption \citep{clip, flamingo, qwen3_tr}. However, recent research suggests that out-of-the-box VLMs remain limited in their application to specialised domains such as ocean and medical sciences \citep{vlm_marine_intelligence, vila_enhancing_vlm_med}. Consequently, in conjunction with more traditional convolutional neural networks (CNNs) \citep{neurocogitron_fukushim_1980, resnet}, vision transformers (ViTs) \citep{vit} remain the standard for classification tasks in these research areas \citep{biigle, reefnet, benthicnet, vits_med_img}. Vision neural networks tend to work through the extraction of low-level latent variables (latents), such as colours, textures, and shapes \citep{visualize_cnns, imagenet_cnns_texture_bias, do_vits_see_like_cnns, why_should_i_trust_you}, resembling the hierarchical visual cortex \citep{neurocogitron_fukushim_1980}. However, these latents lack the generalisable reasoning characteristics that humans use to identify objects. The lack of reasoning in visual tasks within specialised domains therefore remains a research gap. This motivation is illustrated in \autoref{fig:ghiml_motivation}.

\input{figures/fig1_ghiml_motivation/fig1_ghiml_motivation}

Our application domain, benthic (seafloor) imagery classification, supports ecological and economic monitoring of maritime environments under increasing anthropogenic pressure \citep{atlantic_canada_marine_debris}, at a data volume that far exceeds the capacity of manual analysis.

Benthic image classification is often a hierarchical multi-label (HML) task \citep{fathomnet, benthicnet}, meaning a single image may contain multiple organism subjects or substrate types, each situated within a taxonomic or visually derived hierarchy that spans broad morphological categories down to fine-grained species-level distinctions. This combination of multi-label outputs and structured hierarchical dependencies poses challenges that extend beyond standard image classification \citep{coherent_hml, wehrmann_hml}. The problem is further compounded by substantial visual variability in seafloor imagery, arising from inconsistent lighting conditions, colour casts, and diverse substrate appearances, all of which undermine the robustness of learned representations \citep{benthicnet, fathomnet, reefnet, coralnet, biigle}. Moreover, rare classes, often those of the greatest ecological interest, remain poorly understood and difficult to detect by existing HML methods, as limited training examples prevent models from confidently learning discriminative features \citep{hml_missing_info_benthicnet}. Collectively, these factors make benthic image classification a particularly demanding domain in which current approaches struggle to achieve the robustness and reliability that ecological monitoring demands.

In an effort to improve robustness in image detection for this domain, we propose \textbf{G}uided \textbf{Hi}erarchical \textbf{M}ulti-\textbf{L}abel (GHiML) learning, an approach that seeks to unite the general knowledge contained in large VLMs with the specialised training of ViTs. GHiML consists of four steps: (1) identifying high-level latents relevant to reasoning about the domain, data points, and their annotations, (2) generating sample-relevant captions for extracted latents, (3) cross-attention fusion between image embeddings and generated textual embeddings for downstream classification, and (4) training a gated mixture of all three modalities across classes (hierarchical nodes). Our hypothesis is that the relationships between tokens encoded in a sufficiently trained VLM contain useful oceanographic connections beyond what low-level visual latents represent. Conditioning on those latents then helps to elicit this stored knowledge as a sample-specific textual signal.

Additionally, we propose an evaluation score for HML predictions, based on graph traversal distance and comparisons of multi-label cardinality (the degree of ``multi-labelled-ness'') \citep{multi-label_survey} between prediction and annotation. We refer to it as the inverted HML score (\(\dinv\)), a bounded similarity score ranging from 0 to 1, where higher values indicate better predictions. In practice, \(\dinv\) is reported alongside the node-wise (macro-averaged) \(F_1\) score \citep{hml_missing_info_benthicnet, hml_econ_lit}. The two evaluate different objects, with \(F_1\) scoring per-node detection and \(\dinv\) scoring the predicted label set as a whole, and we do not treat either as a substitute for the other.

Our main contributions are therefore:
\begin{enumerate}
    \item A use-inspired pipeline that elicits domain knowledge from a VLM as a sample-specific textual signal and integrates it with an existing HML framework, rather than a new HML method in itself;
    \item \(\dinv\), an evaluation score for HML predictions that accounts for both hierarchical proximity and prediction cardinality; and
    \item An analysis of what the elicited text contributes, which we find acts as a coarse prior over regions of the hierarchy rather than as a discriminative signal within them.
\end{enumerate}

\section{Background}

\subsection{Hierarchical Multi-Label Learning}
In HML, annotations may be envisioned as a path starting from a root and terminating at a node. Multi-label implies that these annotations can consist of any combination of viable paths. The viability of an annotation is determined by the hierarchical constraint, that is \(p_\text{parent} \ge p_{\text{child}}\). To enforce this condition during training, we apply Coherent Hierarchical Multi-Label Classification Neural Networks (C-HMCNN) \citep{coherent_hml}, which has established itself as the predominant framework for constrained HML classification and remains a robust standard \citep{improving_detection_rare_nodes, hml_survey}. C-HMCNN converts the hierarchy of \(|\mathcal{V}|\) nodes (\(\mathcal{V}\) being the set of nodes) into an adjacency matrix (\(\textbf{A} \in \mathbb{R}^{|\mathcal{V}|\times |\mathcal{V}|}\)) that encodes ancestor-descendant relationships, then applies a maximum operation over each node's sub-hierarchy to ensure no child exceeds its parent's predicted probability. This constraint mechanism (\(f_\text{CM}\)) is insufficient alone, and a modified binary cross-entropy (\(\text{BCE}\)) loss is required to ensure gradients point in the correct direction:
\begin{equation}
    \mathcal{L}_{\text{MC}} = \text{BCE}\!\Bigg(
        \underbrace{
            \overbrace{
                (1 - \mathbf{Y})\, f_{\text{CM}}\!\big(\theta(\mathbf{X}),\, \mathbf{A}\big)
            }^{\text{Negative-node contribution}}
            \;+\;
            \overbrace{
                f_{\text{CM}}\!\big(\mathbf{Y}\theta(\mathbf{X}),\, \mathbf{A}\big)
            }^{\text{Positive-node contribution}}
        }_{\text{Full constrained output}},
        \;\; \mathbf{Y}
    \Bigg);
    \label{eq:mcloss}
\end{equation}
where \(\mathcal{L}_{\text{MC}}\) is referred to as the \textit{max constraint loss} (MCLoss) \citep{coherent_hml, improving_detection_rare_nodes}, \(\mathbf{X} \in \mathbb{R}^{n_\text{b} \times d}\) is the input of batch size \(n_\text{b}\) and feature size \(d\), \(\theta\) is the model which maps from \(d\) to \(|\mathcal{V}|\), and \(\mathbf{Y} \in \mathbb{R}^{n_\text{b} \times |\mathcal{V}|}\) is the annotation in binary representation. Finally, the \(\mathbf{A}\) in \autoref{eq:mcloss} has been expanded to be in \(\mathbb{R}^{n_\text{b} \times |\mathcal{V}| \times |\mathcal{V}|}\). Most notably, the positive-node contribution must be filtered by the positive annotations prior to constraint to remove the effect of negatively-annotated descendants on their ancestors.

\section{Methods}
\label{sec:methods}
Our approach is generalised to HML annotations which are not obligated to reach a leaf node (partial-depth). Technical details required for reproduction are provided in 
\autoref{app:methods_technicals}.
\input{figures/fig2_ghiml_captioning/fig2_ghiml_captioning}

\subsection{Extracting Useful Latents}
\label{sec:extracting_latents}

The training partition of an image dataset is supplied to a VLM with an appropriate prompt, as shown in the first part of \autoref{fig:ghiml_captioning}. Formally, using terms from \autoref{fig:ghiml_motivation}, we sample a set of conceptual latents
\(
    \mathbf{Z}_2 \sim p(\mathbf{Z}_2 \mid \mathbf{X}, \mathbf{Y};\, \boldsymbol{\phi})
\),
where the model \(\boldsymbol{\phi}\) must first leverage low-level visual latents \(\mathbf{Z}_1\). The variables in \(\mathbf{Z}_2\) should not resemble those we might expect from \(\mathbf{Z}_1\). The goal is not to generate a description of the image and introduce redundant information for a visual encoder. Instead, the usefulness of \(\mathbf{Z}_2\) is predicated on its ability to bias the model towards conceptual knowledge acquired through pretraining on large-scale language data.

With a sufficiently large and diverse dataset, this procedure can return thousands of candidate variables, many of which are redundant or weakly informative. We therefore refine the proposed set in four steps, specified in full in \autoref{app:latent_extraction_details}. (1) We compute the empirical distribution \(p(v)\) over nodes from the training annotations. (2) Since samples may be multi-labelled, an EM procedure resolves the ambiguous association between each proposed variable \(z_{2,j}\) and individual nodes, yielding a distribution \(q_j(v)\) over nodes. (3) Each \(q_j\) is smoothed towards the prior, and candidates are scored by \(D_{\mathrm{KL}}(\tilde{q}_j \,\|\, p)\); variables with high divergence concentrate their annotation mass on rarer nodes and are hypothesised to carry the most discriminative information beyond the prior, and we retain only those scoring \(\geq 1\). (4) After deduplication, a uniform per-node budget \(B\) selects the top-\(B\) candidates by KL score for each leaf node, as providing smaller VLMs with a large quantity of keywords can introduce anchoring bias \citep{llms_easily_distracted, llms_do_in-context_learning_differently}. The final variable set typically comprises 15-20 variables in our runs.

\subsection{Latent Captioning}
We refer again to \autoref{fig:ghiml_captioning}. This step consists of supplying the processed set of variables to the VLM for captioning. In addition to captioning for each variable, the model is prompted to assign a binary score for the relevance of the variable to each sample. This is done for analysis purposes, and is inspired by \citet{coat_discovery}. A quality check is performed to ensure that all variables are accounted for in the captions of each sample.

\subsection{Models and Architecture}
\input{figures/fig3_architectures/fig3_architectures}

We refer to \autoref{fig:architecture} for an overview of the models and training pipeline we use.
Images are embedded with a DINOv3 ViT-H backbone \citep{dinov3}, Qwen3-VL-8B-Instruct is used for variable proposal and captioning, and Qwen3-Embedding-8B for embedding captions \citep{qwen3_tr}. These models were chosen for their open-source availability, reliability, and ability to fit on consumer-grade GPUs accessible to domain-specialised research labs.

To summarise our models, we train self-attention-based pooling mechanisms to transform sequential image and text embeddings into dense vectors. These unimodal models serve as baselines and reveal the overlap between textual and visual performance, specifically the degree to which each modality carries redundant or complementary information. A third cross-attention mechanism is trained with image as the query modality and text as the key-value modality\footnote{In preliminary testing, image querying text performed stronger than the reverse.}. Finally, we train a router \citep{adaptive_moe_local} that takes as input the concatenated pooled embeddings and outputs softmax weights over the three modalities for each node, so that the contribution of each modality may vary by node and by sample. We run two versions of this router: one that loads frozen pretrained pooling layers from our unimodal runs and trains the classifiers and router mechanism (Router), and one with all parameters trained from scratch (Router\textsuperscript{\dag}).

Every row of \autoref{tab:results} uses the same hierarchical classification head, an implementation of C-HMCNN \citep{coherent_hml} with the max-based constraint \(f_\text{CM}\) and its accompanying loss, and the same per-node thresholding procedure (\autoref{sec:thresholding}). The rows therefore differ only in which modalities are available and how they are combined, so C-HMCNN is held constant as a control rather than compared against.

\subsection{Inverted HML Score \texorpdfstring{\(\dinv\)}{d-inv}}

Evaluating HML predictions requires a score that captures both \textit{closeness} between prediction and ground truth and \textit{multi-label cardinality}, penalising predictions that are deficient or excessive, given that real-world annotations may be incomplete \citep{benthicnet, fathomnet}. Existing scores do not fully satisfy these requirements. Node-wise \(F_1\) penalises cardinality errors but forfeits hierarchical proximity, particularly at standard \(0.5\) thresholding, where deeper nodes frequently score zero regardless of prediction closeness \citep{improving_detection_rare_nodes}. Micro-averaged AP \citep{coherent_hml, wehrmann_hml} pools across all nodes, so evaluation in sparse hierarchies lets easily ranked negatives dominate, producing overly optimistic scores. A comparison of these scores is presented in \autoref{app:hml_metrics_study}.

We address these gaps with \(\dinv\), which directly exploits hierarchical distance and explicitly accounts for multi-label cardinality (with user-stipulated flexibility). Because it maps graph traversal distance onto a normalised \([0,1]\) scale, the score is directly interpretable: a value of one indicates a perfect prediction, while intermediate values communicate precisely how far the prediction sits in terms of the hierarchy's own structure, and how far its cardinality departs from the annotation. We define:
\begin{equation}
    d_{\text{inv}} = s_{\max}\,\gamma,
\end{equation}
where \(s_{\max}\) encapsulates both normalisation and ``identification'' (determining which multi-label component in the prediction corresponds to which component in the ground truth). Both \(\hat{\mathcal{Y}}\) and \(\mathcal{Y}\) are ancestor-closed by construction, and \(\dinv\) is computed over their \emph{endpoints}: the active nodes carrying no further active descendants. Ancestors are therefore not scored twice, and a prediction is evaluated by where its most specific claims fall. For normalisation, we map the prediction distance \(d_{\text{pred}}^{(ij)}\) between ground-truth endpoint \(i\) and predicted endpoint \(j\) onto a \([0,1]\) scale via \(\left[\mathbf{D}_{\text{inv}}\right]_{ij} = (d_{\max}^{(i)}-d_{\text{pred}}^{(ij)})/d_{\max}^{(i)}\), where \(d_{\max}^{(i)}\) is the eccentricity of ground-truth endpoint \(i\), i.e.\ the greatest tree distance from it to any other node in the hierarchy\footnote{Under unit edge weights.}. For identification, we construct a matrix \(\mathbf{D}_{\text{inv}} \in \mathbb{R}^{n_{\text{truth}}\times n_{\text{pred}}}\) of pairwise similarity scores for all component combinations, and solve the resulting rectangular assignment problem using the Jonker--Volgenant algorithm \citep{2-d_rect_assign}. Writing \(f_{\text{JV}}(\hat{\mathcal{Y}}, \mathcal{Y}; \mathbf{D}_{\text{inv}})\) for the total similarity of the optimal assignment, we normalise by the number of ground-truth components to obtain
\begin{equation}
    s_{\max} = \frac{1}{n_{\text{truth}}}\,
    f_{\text{JV}}(\hat{\mathcal{Y}}, \mathcal{Y};\, \mathbf{D}_{\text{inv}}),
\end{equation}
the optimistically matched combination score between prediction
(\(\hat{\mathcal{Y}}\)) and ground truth (\(\mathcal{Y}\)). The cardinality factor \(\gamma\) penalises excessive predictions, an important consideration in HML settings. Defining the number of excessive predictions as \(n_{\text{E}} = n_{\text{pred}}-n_{\text{truth}}\) and the maximum number of excessive predictions as \(n_{\text{ME}} = |\mathcal{V}| -n_{\text{truth}}\), we model \(\gamma\) with exponential decay:
\begin{equation}
    \label{eq:exponential_decay_form}
    \gamma(n_{\text{E}};\,k_{\text{P}},\,n_{\text{ME}}) = 
    \begin{cases} 
        a\, e^{-b\, n_{\text{E}}} + c & \text{if } k_{\text{P}} \neq k^{*}_{\text{P}}, \\[4pt]
        1-\dfrac{n_{\text{E}}}{n_{\text{ME}}} & \text{otherwise},
    \end{cases}
\end{equation}
where \(k_{\text{P}} \in (0,1)\) governs the penalty incurred by a single excess prediction, \(k^{*}_{\text{P}} = (n_{\text{ME}}-1)/n_{\text{ME}}\) is the critical value at which the decay reduces to the linear case, and the parameters \(a = (1-e^{-bn_{\text{ME}}})^{-1}\) and \(c = 1-a\) are determined by boundary conditions \(\gamma(0)=1\) and \(\gamma(n_{\text{ME}})=0\), with \(b\) obtained as the unique non-trivial root of \((k_{\text{P}}-1)e^{-bn_{\text{ME}}}+e^{-b}-k_{\text{P}}=0\) (formal proofs of existence and uniqueness are provided in \autoref{app:inv_hml_dist_proofs}). Edge cases (disjoint roots, empty and deficient predictions) are defined in \autoref{app:inv_hml_dist_proofs}.

As \(\dinv\) is defined per sample, we report it macro-averaged across nodes, as with \(F_1\): each node receives the mean \(\dinv\) of the test samples annotated with it, and these node scores are averaged with equal weight. We set \(k_\text{P} = 0.5\), as a significant contemporary challenge in the benthic domain is the possibility of incomplete annotations \citep{fathomnet-2026}. Consequently, the annotation cardinalities cannot be fully trusted.

\subsection{Optimal \texorpdfstring{\(F_1\)}{F1} Thresholding}
\label{sec:thresholding}
Deeper and rare nodes are poorly detected in HML \citep{hml_imbalance_hros, optimal_thresholding_multi-label_f1}, possibly in part due to poor calibration \citep{hml_missing_info_benthicnet}. Following \citet{optimal_thresholding_multi-label_f1}, we grid-search per-node thresholds maximising \(F_1\) (\(0.01\) to \(0.99\) in \(0.01\) increments, starting from \(0.50\)), subject to the hierarchical constraint \(t_{\text{parent}} \leq \min_{v \in \mathcal{V}_{\text{d}}} t_v\), where \(\mathcal{V}_{\text{d}}\) denotes the set of descendants. Thresholds are fitted on training predictions only, and the procedure is universally applied to every model reported.


\section{Experiments}
\label{sec:experiments}

\subsection{Datasets}
\label{subsec:datasets}
We evaluate on three datasets, described in full in \autoref{app:datasets}. BenthicNet \citep{benthicnet} provides 188,679 annotated images labelled under the CATAMI scheme \citep{catami}, from which we form two overlapping tasks, substrate (BN-Sub, 23 nodes) and biota (BN-Bio, 293 nodes). FathomNet (2025) (FN) \citepalias{fathomnet-2025} is a benthic biota competition dataset of 9,305 full images, for which we extracted a 192-node taxonomic hierarchy from WoRMS \citep{worms}, making the problem HML (full-depth). For both, the innate test partitions serve as out-of-domain (OOD) test sets, and we additionally construct in-domain (ID) test sets from a random 20\% of the data. These partitions are ``OOD'' because BenthicNet's test set is geospatially partitioned and FathomNet's competition test set was hand-selected. Lastly, iNaturalist (iNat) \citep{inat} spans terrestrial fauna, flora, and fungi with single-label annotations over a hierarchy of over 16,000 nodes; we train on the 500,000 training samples of iNat (mini) and test on its 100,000 validation samples, replacing MCLoss with a softmax over the 10,000 endpoints, marginalised up the hierarchy. Since each prediction and annotation consists of exactly one endpoint, \(n_\text{E} = 0\) and \(\gamma \equiv 1\), so \(\dinv\) reduces to normalised traversal similarity on this dataset.

\subsection{Results and Discussion}
\label{subsec:results_and_discussion}
\input{tables/table1_benthicnet}
\input{tables/table1-5_inat}

As shown in \autoref{tab:results}, the router models achieve the strongest performance in nearly all settings. The ranking across methods is consistent for each metric across all random seeds. Although text underperforms vision on \(F_1\) at nearly every node, the two metrics disagree in a specific way: at 75 of the 171 BN-Bio nodes with sufficient support, text scores higher on \(\dinv\) while failing to identify the node on \(F_1\), against 6 nodes where the reverse holds. This indicates that text contributes generalisable semantic information about where predictions belong in the hierarchy, rather than finer discrimination between nodes.

One contributing factor is that several of the selected variables encode functional traits, such as mobility, attachment strategy, and predation mode, that are particularly informative for sessile organisms. Algae and sponges, for instance, share many such traits, and the resulting captions place them in close proximity within the hierarchy even when exact classification fails.

\autoref{tab:inat_results} reports the same comparison on iNat, where the hierarchy is roughly \(55\) times larger and annotations are single-label. The ordering is unchanged: both routers exceed the image branch on both metrics at every seed, and text alone remains the weakest by a wide margin on \(F_1\) while retaining half of the image branch's \(\dinv\). Because \(\dinv\) has no cardinality term on iNat, this is a direct measure of proximity. A text stream with poor \(F_1\) on its own still places predictions within reach of the target.

We next ask whether text and vision contribute distinct information, and whether the router successfully exploits both. Among the top-performing nodes for text, there is low overlap between the two modalities, and the router frequently surpasses vision while retaining a margin of advantage comparable to that of text alone (\autoref{app:bn-bio_case_study}). There, we also study the relevant variables and their correlations with text advantage, as well as with individual nodes. We observe that the proposed variables provide strong contextual information that benefits classification, even when they do not directly improve prediction of the original nodes with which they were associated via KL divergence.

This last point warrants emphasis. Although variables were assigned to modal nodes, the captions they generate encode a far richer set of combinations. The variable \texttt{ambush}, for example, was initially selected for its association with sea stars. In practice, it also describes rocky substrates (habitats in which ambush predators tend to occur), detects lobsters (themselves ambush predators), and identifies organisms or environments facilitating camouflage consistent with ambush strategy. A single functional concept thus provides signal across multiple nodes. This is consistent with considering ecological function and context rather than relying solely on low-level visual latents such as colour and texture.

\subsection{Text as a Sub-Tree Prior}
\label{subsec:subtree_prior}

Beyond the trained branches, we also examine the embeddings themselves. We fit two \(k\)-NN classifiers over the frozen image and text embeddings, mean-pooled across the sequence dimension, and evaluate them on the same samples by cross-validation (\autoref{app:caption_probe}). No trained component of our pipeline is used, so any complementarity observed is a property of the representations rather than of our architecture. Were the text embedding correct only on samples that the image embedding already correctly classifies, it would be strictly redundant and the independence ratio \(P(\text{text correct}\mid\text{image wrong})\,/\,P(\text{text correct})\) would be zero. Across the BN and iNat datasets, \(3.5\)--\(10.8\%\) of samples across hierarchy levels are classified correctly by the text embedding but not by the image embedding. A more detailed \(k\)-NN comparison against generic captioning (image description) embeddings in BN-Bio (\autoref{app:caption_probe}) demonstrates that our extracted keywords help reduce redundancy with images precisely at the clade levels which benefit classification the most.

\input{tables/table6_mean_replacement}

In \autoref{tab:mean_replacement}, a mean embedding vector is supplied to the models in lieu of the original text embedding, such that no per-sample textual information survives. On BN-Bio the degradation occurs only for \(\dinv\), while \(F_1\) is unaffected. On iNat both metrics degrade. BN-Sub, the smallest hierarchy, is unaffected in either metric. The text signal therefore appears to matter more for larger and more complex hierarchies. The two biotic datasets differ in which metrics respond, and we attribute this to their label structure. Under multi-label annotations, the clade signal appears to suppress spurious subtree activations, which is expressed through prediction cardinality rather than hierarchical placement (\autoref{app:decomposition}). However, iNat's single-label annotations foreclose that channel. Replacing the text with scale-matched Gaussian noise degrades performance below the image-only baseline on three of the four datasets, indicating that the gains do not follow from the added dimensionality alone (\autoref{app:noise}). 

We also report on zero-shot VLM classification, which produced unscorable results. A zero-shot VLM cannot produce scheme-conformant predictions for this task without conditioning, and even with scheme-adherent conditioning, it over-predicts substantially (\autoref{app:zeroshot}). Additionally, we test a descriptor-matching baseline, which recovers roughly half the node \(F_1\) of a trained image branch and under a tenth of its \(\dinv\), with descriptors reused heavily between sibling nodes (\autoref{app:dclip}). Many benthic images are homogeneous in colour, texture and tone despite differing content, so generic appearance-based descriptions tend to be interchangeable across nodes. Finally, we analyse the router's gate weights on BN-Bio and confirm that they remain open to text (\autoref{app:gate}).

\section{Related Works}
\label{sec:related_works}
Our work is influenced by the causal discovery literature, most notably COAT \citep{coat_discovery}. COAT employs LLMs to discover a phenomenon's Markov Blanket by proposing candidate variables and annotating each sample with a trinary encoding per variable (relevant, irrelevant, or inversely relevant). COAT demonstrates that LLMs can actively mediate between unstructured observations and structured representations. Our pipeline differs in that we target hierarchical multi-label classification, select variables per hierarchy node by KL divergence, and have a VLM score each variable with a brief visual justification. The resulting captions are embedded as dense text and fused with the image representation through a learned per-node router, so the variables serve as contextual signal for a vision model and are never used as discrete features for causal analysis.

Closer in line with our motivation are diffusion-based augmentation strategies \citep{semantic_augmentation_in_images, sgid, effective_data_augmentation_diffusion, seed_select}, which leverage the visual priors of large-scale generative models to augment downstream classification. SGID \citep{sgid} uses BLIP-generated captions \citep{blip} and CLIP-based selection \citep{clip} to guide Stable Diffusion toward class-consistent samples, DA-Fusion \citep{effective_data_augmentation_diffusion} semantically augments real images via Stable Diffusion, and SeedSelect \citep{seed_select} optimises the initial noise seed to improve rare class generation. However, these approaches are largely inapplicable to benthic habitat classification, where many taxa remain underrepresented in the web-scale data on which diffusion models are trained \citep{dreamsea}. Instead, much of our knowledge of benthic taxa and habitat is encoded textually, motivating our alternative strategy of extracting higher-level semantic information from a VLM rather than synthesising novel training samples.

Within the HML literature, research into improving detection and addressing class imbalance remains sparse, with existing methods broadly employing either resampling techniques \citep{hml_imbalance_hros} or weighted objectives \citep{improving_detection_rare_nodes}. Our work enhances node detection broadly by leveraging the general knowledge of an external foundation model.

Lastly, we compare \(\dinv\) with the hierarchical \(F_1\) of \citet{hml_econ_lit} and with CoPHE \citep{cophe} in \autoref{app:sub:inv_hml_dist_vs_hf1}. We find \(hF_1\) comparatively lenient toward excessive prediction cardinality, since excess predictions that share correct ancestors inflate its ancestor-expanded sets only marginally, whereas CoPHE, which addresses this leniency, does not measure how far a prediction sits from its target and requires a well-defined depth for every node.


\section{Conclusion}
We have presented GHiML, a pipeline for extracting and leveraging higher-level semantic knowledge from VLMs to augment hierarchical multi-label classification in specialised visual domains. Through a structured variable proposal, EM-based filtering, and sample-conditioned captioning procedure, we demonstrated that a VLM can be positioned to generate complementary contextual information to standard visual features. The gated router architecture exploits this information on a per-node basis, consistently outperforming both unimodal baselines and the cross-attention fusion model across two benthic datasets under both in-domain and out-of-domain considerations, and on iNat.

We additionally introduced \(\dinv\), a metric that grounds HML evaluation in the structure of the hierarchy itself. By combining graph traversal distance with an explicit cardinality penalty governed by a single interpretable parameter, \(\dinv\) addresses gaps left by existing methods, and offers an intuitive understanding of model performance. Limitations of the work are discussed in greater detail in \autoref{app:limitations}.

\subsection*{Acknowledgements}
This research was supported by the Ocean Frontier Institute (OFI) through an award from the Canada First Research Excellence Fund, and as part of the Benthic Ecosystem Mapping and Engagement (BEcoME) project. Computational resources were provided by the Digital Research Alliance of Canada. We thank Dr.\ Mariano Maisonnave for helpful discussions.

\bibliography{main}
\bibliographystyle{plainnat}


\appendix

\section{Technical Details}
\label{app:methods_technicals}
\input{tables/table2_task_diagnostics}
\input{tables/table3_training_diagnostics}
\input{tables/table4_hyperparameters}

\subsection{Attention-Based Pooling}
We consider two architectures for producing a fixed-dimensional representation from variable-length token sequences, both sharing a common structure consisting of a contextual encoding stage followed by learned query attention pooling. In the self-attention pooling variant, given an input sequence \(\mathbf{X} \in \mathbb{R}^{n_\text{b} \times T \times d}\) of batch size \(n_\text{b}\), sequence length \(T\), and embedding dimension \(d\), we first apply multi-headed (four heads) self-attention with residual connections and layer normalisation \citep{layernorm}, followed by a position-wise feed-forward network, yielding a contextually enriched sequence \(\tilde{\mathbf{X}} \in \mathbb{R}^{n_\text{b} \times T \times d}\). In the cross-attention fusion variant, two distinct token sequences are available, a query sequence \(\mathbf{X}_q \in \mathbb{R}^{n_\text{b} \times N \times d}\) (image embeddings) and a key-value sequence \(\mathbf{X}_{kv} \in \mathbb{R}^{n_\text{b} \times M \times d}\) (text embeddings), and we replace the self-attention encoder with an analogous cross-attention layer in which \(\mathbf{X}_q\) attends over \(\mathbf{X}_{kv}\), followed by the same residual connections, layer normalisation, and feed-forward network, producing a fused sequence \(\tilde{\mathbf{X}}_q \in \mathbb{R}^{n_\text{b} \times N \times d}\) that integrates information from \(\mathbf{X}_{kv}\) into the positions of \(\mathbf{X}_q\). In both cases, the final pooling stage is identical: a single learnable query parameter \(\mathbf{q} \in \mathbb{R}^d\) attends over the encoded sequence via multi-head cross-attention, followed by layer normalisation, yielding a fixed-dimensional representation \(\mathbf{z} \in \mathbb{R}^d\). Padding masks are propagated through all attention stages to ensure that padded positions contribute neither to contextual encoding nor to the final pooled representation. The resulting vector \(\mathbf{z}\) is subsequently passed to an HML constrained feed-forward classifier.

\subsection{Gated Router}
Given image, text, and cross-modal streams, we extract pooled feature vectors \(\mathbf{z}_{\text{img}}, \mathbf{z}_{\text{txt}}, \mathbf{z}_{\text{cross}} \in \mathbb{R}^{d}\) and pass each through a dedicated classification head for each stream. This yields per-stream logit matrices stacked into a tensor of shape \((n_\text{b}, 3, |\mathcal{V}|)\). A gating network ingests the concatenated features \([\mathbf{z}_{\text{img}};\, \mathbf{z}_{\text{txt}};\, \mathbf{z}_{\text{cross}}] \in \mathbb{R}^{3d}\) and produces per-sample, per-node softmax weights over the three modes. The gate applies layer normalisation \citep{layernorm}, a linear projection from \(3d\) to \(d\), GELU activation with dropout, and a final linear layer from \(d\) to \(3|\mathcal{V}|\), reshaped to \((n_\text{b}, |\mathcal{V}|, 3)\) and softmax-normalised along the last axis. The fused logit for node \(v\) is then
\begin{equation}
    l_v = \sum_{s \in \{\text{img},\, \text{txt},\, \text{cross}\}} g_v^{(s)} \cdot l_v^{(s)},
\end{equation}
where \(g_v^{(s)}\) is the gate weight and \(l_v^{(s)}\) is the stream logit. Sigmoid activation and the hierarchical constraint from C-HMCNN are applied after blending.

\subsection{Diagnostics}
All diagnostic experiments were conducted on a single NVIDIA GeForce RTX 4090 GPU. Computational costs for VLM inference are reported in \autoref{tab:task_diagnostics}, and training diagnostics are provided in \autoref{tab:training_diagnostics}. We set the embedding dimension \(d\) to 1024 for Qwen3-Embedding-8B (which uses Matryoshka representations \citep{matryoshka_rep_learning}), matching the output dimensionality of DINOv3's ViT-H backbone. Each HML classifier head consists of a 3-layer MLP with hidden dimension 1024, GELU activations \citep{gelu}, and dropout of 0.7, mapping from the 1024-dimensional embedding space to the label space \(\mathbb{R}^{|\mathcal{V}|}\). We optimise with Adam \citep{adam} using a weight decay of \(1\mathrm{e}{-4}\) across all experiments. Learning rates and training epochs are reported in \autoref{tab:training_hyperparams}. Training epochs were selected based on validation loss curves. BenthicNet's larger dataset was less sensitive to overfitting, permitting 35 epochs to convergence, whereas FathomNet required more careful tuning to mitigate overfitting.

\section{Captioning Embeddings}
\label{app:caption_probe}
\input{tables/table5_embeddings_analysis}
\input{tables/table11_caption_embeddings_compare}

\autoref{tab:emb_probe} reports the frozen-embedding probe summarised in \autoref{subsec:subtree_prior} by hierarchy level. iNaturalist level~1 is excluded because image accuracy there reaches \(99.6\%\), leaving a conditioning set of only \({\sim}89\) samples. We then ask whether this complementarity depends on the captions being built on the extracted variables.

We compare the elicited captions against four alternatives generated by the same VLM for the same images: a plain caption, a visual description restricted to shapes, colours and textures, a description of the organisms and seafloor, and the same description additionally asked to relate the image to the extracted variables (\autoref{app:prompts}). At the coarsest level of the hierarchy, the captions built on the extracted variables are markedly less redundant with the image than the generic captions (\autoref{tab:caption_probe}). The description and keyword prose prompts differ only by the reference to the extracted variables, and this addition alone accounts for most of the gap. At deeper levels the five sources are comparable; however, we note that these deeper levels are where we expect text to be less useful in general.

\section{Dataset Details}
\label{app:datasets}

\subsection{BenthicNet}
BenthicNet \citep{benthicnet} is one of the largest publicly available ocean science datasets, comprising over 11 million benthic images aggregated from 2,058 constituent datasets spanning a wide variety of geographic regions, depth ranges, and imaging platforms, sourced from research groups and government agencies around the world. Its label taxonomy follows the Collaborative and Annotation Tools for Analysis of Marine Imagery (CATAMI) visual classification scheme \citep{catami}, and is unique in that annotations span both substrate and biota categories, facilitating both habitat and organism classification. We draw from it its full set of annotated imagery (188,679 images), forming two overlapping subsets for substrate and biota (BN-Sub and BN-Bio, respectively), which we treat as independent classification tasks. BN-Sub contains 23 nodes while BN-Bio contains 293. BenthicNet's test partition is geospatially subsampled, meaning test images originate from regions unseen during training. We refer to this innate partition as the out-of-domain (OOD) test set, and additionally construct an in-domain (ID) test set through random stratified sampling (of 20\% of the data) for evaluation under more conventional conditions. The combination of global geographic coverage and diversity across both biota and substrate categories makes BenthicNet a suitable testbed for HML methods that must generalise across heterogeneous visual conditions.

\subsection{FathomNet (2025)}

FathomNet (2025) (FN) \citepalias{fathomnet-2025} is a benthic biota classification competition dataset consisting of 9,305 full images and 24,487 regions-of-interest (ROIs). Full images may contain multiple ROIs. The dataset is not inherently hierarchical, despite using taxonomic annotations. Furthermore, it has been curated to be balanced across classes. We extracted a hierarchy of 192 nodes spanning phylum, class, order, family, genus, and species\footnote{For some annotations, these levels were replaced with a sub-level alternative where necessary.} from the World Register of Marine Species (WoRMS) \citep{worms}. Subsequently, the classes in the dataset become leaf nodes, and the problem becomes HML (full-depth). While earlier imagery from FathomNet was contained in BenthicNet, the 2025 competition dataset is novel with a carefully curated OOD test set (\citealp{benthicnet}; \citetalias{fathomnet-2025}). As with BenthicNet, we use both the innate test partition and a random partition (of 20\%) for OOD and ID evaluation respectively.

\subsection{iNaturalist}
\label{subsec:inat}
To demonstrate the generalisability of the proposed approach, experiments are also performed on the iNaturalist (iNat) \citep{inat} dataset. The iNat dataset encompasses a broader domain than BenthicNet, capturing terrestrial fauna, flora, and fungi. Additionally, it presents a distinct structural classification challenge. While the dataset inherently contains 10,000 distinct species classes, its biological taxonomic hierarchy allows for the construction of a hierarchical classification problem spanning over 16,000 nodes, characterised by mutually exclusive, single-label assignments.
In adapting our framework for iNat, we replace MCLoss with a softmax over the 10,000 endpoints, marginalised up the hierarchy, so that each node receives the summed probability of its descendants and the hierarchical structure is preserved. Training occurs on the 500,000 samples from iNat (mini), and the 100,000 validation samples act as a test set (since the true test partition is not publicly available).

\section{Prompt Templates}
\label{app:prompts}

We use two prompt templates for variable proposal and captioning, and four further templates for the caption comparison in \autoref{app:caption_probe}. Where applicable, templates are parameterised by task-specific variables denoted with braces.
  
\subsection{Variable Proposal Prompt}
\label{app:prompt_proposal}

\begin{tcolorbox}[colback=gray!5, colframe=gray!50, title=Variable Proposal Prompt, fontupper=\small\ttfamily]
You are a marine benthic ecologist.\\
This seafloor image has the following annotations: \{annotations\}.\\
Use these annotations to guide which ecological properties would be most diagnostic for identifying the annotated categories.\\[4pt]
Task: Infer functional ecological properties that help distinguish between different \{task\}.\\[4pt]
Go beyond what a simple visual description would capture --- use ecological reasoning to infer properties from what you observe. Focus especially on diagnostic properties where ecological and world knowledge would help resolve highly specific or rare categories.\\[4pt]
Do NOT return simple visual descriptions (e.g., color, shape, texture). Do not restate the annotation categories themselves. Return up to 10 of the most diagnostic properties as 1--3 lowercase words each, separated by commas. Return nothing else.
\end{tcolorbox}

\subsection{Captioning Prompt}
\label{app:prompt_scoring}

\begin{tcolorbox}[colback=gray!5, colframe=gray!50, title=Captioning Prompt, fontupper=\small\ttfamily]
You are a marine benthic ecologist analyzing a seafloor photograph.\\[4pt]
Task: For each ecological property below, carefully reason about whether it is supported or contradicted by visible evidence of the \{task\} in this image. Connect what you SEE to what you KNOW about benthic ecology. Score each property:\\
\quad 1 = visually supported\\
\quad 0 = not applicable or visually contradicted\\[4pt]
Properties:\\
\{variables\}\\[4pt]
Format:\\
\quad [property\_term]: [brief reasoning and justification (one sentence, \textasciitilde5 words)] [score value]\\
Adhere strictly to the format. Output one property per line.
\end{tcolorbox}

\noindent Here, \texttt{\{annotations\}} denotes the ground-truth categories expressed in natural language, 
\texttt{\{task\}} specifies the ecological domain (e.g., ``biota'' or ``substrate characteristics''), 
and \texttt{\{variables\}} is the alphabetically sorted set of all processed variables for the domain.

The following prompts generate the alternative captions compared in \autoref{app:caption_probe}. All four use the same VLM and decoding settings as the captioning prompt.

\begin{tcolorbox}[colback=gray!5, colframe=gray!50, title=Plain Caption Prompt, fontupper=\small\ttfamily]
Describe the image in a single paragraph of about 150 words.
\end{tcolorbox}

\begin{tcolorbox}[colback=gray!5, colframe=gray!50, title=Visual Caption Prompt, fontupper=\small\ttfamily]
Describe in detail what is directly visible in the image: shapes, colours, textures. Write a single paragraph of about 150 words.
\end{tcolorbox}

\begin{tcolorbox}[colback=gray!5, colframe=gray!50, title=Description Prompt, fontupper=\small\ttfamily]
You are looking at a seafloor photograph.\\[4pt]
Task: Describe in detail what is visible in the image: the organisms present and the seafloor.\\[4pt]
Write a single paragraph of about 150 words. Return nothing else.
\end{tcolorbox}

\begin{tcolorbox}[colback=gray!5, colframe=gray!50, title=Keyword Prose Prompt, fontupper=\small\ttfamily]
You are looking at a seafloor photograph.\\[4pt]
Task: Describe in detail what is visible in the image: the organisms present and the seafloor. Where relevant, relate your description to the following ecological properties: \{variables\}.\\[4pt]
Write a single paragraph of about 150 words. Return nothing else.
\end{tcolorbox}

\section{Latent Extraction Details}
\label{app:latent_extraction_details}

This section provides the complete mathematical specification of the variable extraction pipeline summarised in \autoref{sec:extracting_latents}.

\subsection{Step 1: Node Frequency Prior}

We compute the empirical distribution over nodes from the training annotations:
\begin{equation}
    p(v) = \frac{n_v}{\sum_{v' \in \mathcal{V}} n_{v'}}, \quad \forall \, v \in \mathcal{V}.
\end{equation}
\subsection{Step 2: EM Decomposition}

We define the observed annotations for a candidate \(z_{2,j} \in \mathbf{Z}_2\) as a set of label-count pairs \(\{(\mathcal{A}_k, m_k)\}\), where \(\mathcal{A}_k \subseteq \mathcal{V}\) is a set of co-occurring partial-depth paths within a single sample and \(m_k\) is the number of samples associated with \(z_{2,j}\) that carry annotation set \(\mathcal{A}_k\). We partition these into single-path pairs \(\mathcal{S}_j = \{(\mathcal{A}_k, m_k) : |\mathcal{A}_k| = 1\}\) and multi-path pairs \(\mathcal{M}_j = \{(\mathcal{A}_k, m_k) : |\mathcal{A}_k| > 1\}\). Let \(\mathcal{S}_j(v) = \{k : (\mathcal{A}_k, m_k) \in \mathcal{S}_j,\; v \in \mathcal{A}_k\}\) and \(\mathcal{M}_j(v) = \{k : (\mathcal{A}_k, m_k) \in \mathcal{M}_j,\; v \in \mathcal{A}_k\}\).

Counts are initialised from unambiguous single-path instances:
\begin{equation}
    n_j^{(0)}(v) = \sum_{k \in \mathcal{S}_j(v)} m_k.
\end{equation}
Multi-path counts are then iteratively distributed in proportion to current estimates:
\begin{equation}
    n_j^{(t+1)}(v) = n_j^{(0)}(v) \;+\; 
    \sum_{k \in \mathcal{M}_j(v)} 
    m_k \cdot \frac{n_j^{(t)}(v)}{\sum_{v' \in \mathcal{A}_k} n_j^{(t)}(v')},
\end{equation}
iterating until convergence:
\begin{equation}
    \max_{v \in \mathcal{V}} \left| n_j^{(t+1)}(v) - n_j^{(t)}(v) \right| < \tau.
\end{equation}
The resulting distribution over nodes for each \(z_{2,j}\) is:
\begin{equation}
    q_j(v) = \frac{n_j(v)}{\sum_{v' \in \mathcal{V}} n_j(v')}.
\end{equation}

\subsection{Step 3: Smoothing and KL Scoring}

Before computing divergence, we smooth each \(q_j\) towards the prior \(p\) by adding a pseudo-count volume equivalent to Laplace smoothing, distributed according to \(p\):
\begin{equation}
    \tilde{q}_j(v) = \frac{n_j(v) + |\mathcal{V}| \cdot p(v)}
    {N_j + |\mathcal{V}|},
\end{equation}
where \(N_j = \sum_{v'} n_j(v')\) is the total count mass for \(z_{2,j}\). This smoothing serves as a sample-size-aware regulariser. A variable supported by many annotated instances will have large \(N_j\), rendering the pseudo-counts negligible and preserving the empirical distribution, whereas a variable supported by only a handful of instances will be pulled substantially towards the prior.

A key assumption underlies this scoring procedure. The empirical posterior \(q_j(v)\) is constructed from the same annotation source from which the candidate variables were originally proposed, so high \(q_j(v)\) reflects co-occurrence within the annotation pipeline rather than an independently observed predictive relationship. We assume that this co-occurrence is indicative of genuine discriminative value. In other words, a candidate \(z_{2,j}\) whose annotation mass concentrates on a node \(v\) does indeed capture a visual or semantic property informative for distinguishing \(v\) from other nodes. This assumption may not hold universally, but provides the working basis for using the KL divergence as a proxy for discriminative value. Candidates are then scored by their divergence from the prior,
\begin{equation}
    D_{\mathrm{KL}}\!\big(\tilde{q}_j \;\|\; p\big) = \sum_{v \in \mathcal{V}} \tilde{q}_j(v) \log \frac{\tilde{q}_j(v)}{p(v)},
\end{equation}
and only candidates scoring \(\geq 1\) are retained.

\subsection{Step 4: Deduplication and Selection}

The candidate set emerging from \(D_{\mathrm{KL}}\) filtering undergoes three stages of refinement.

\paragraph{Distribution-based deduplication.}
Each surviving candidate is represented by its normalised distribution \(\tilde{q}_j\). Processing in descending KL order, a candidate is discarded if its cosine similarity with any previously retained distribution meets or exceeds a threshold \(\tau_{\mathrm{dist}}\):
\begin{equation}
    \frac{\tilde{q}_j \cdot \tilde{q}_k}
         {\|\tilde{q}_j\|\;\|\tilde{q}_k\|}
    \;\geq\; \tau_{\mathrm{dist}}.
\end{equation}
We use \(\tau_{\mathrm{dist}} = 0.99\).

\paragraph{Coverage-based deduplication.}
Let \(\mathcal{I}(z_{2,j}) \subseteq \mathbf{X}\) denote the set of training images whose proposed-variable field contains \(z_{2,j}\). Candidates are processed in descending KL order. A candidate is discarded if its image set has Jaccard similarity at or above \(\tau_{\mathrm{cov}}\) with any previously retained candidate:
\begin{equation}
    J\!\big(\mathcal{I}(z_{2,j}),\, \mathcal{I}(z_{2,k})\big)
    = \frac{|\mathcal{I}(z_{2,j}) \cap \mathcal{I}(z_{2,k})|}
           {|\mathcal{I}(z_{2,j}) \cup \mathcal{I}(z_{2,k})|}
    \;\geq\; \tau_{\mathrm{cov}}.
\end{equation}
We use \(\tau_{\mathrm{cov}} = 0.8\).

\paragraph{Per-node top-\(B\) selection.}
Surviving candidates are assigned to their modal node \(v_j^* = \arg\max_{v \in \mathcal{V}} n_j(v)\), inducing per-node groups \(\mathcal{Z}(v) = \{z_{2,j} : v_j^* = v\}\). A uniform budget \(B\) is applied to leaf nodes \(\mathcal{V}_L \subseteq \mathcal{V}\), prioritising rarity. For each leaf node \(v\) with \(\mathcal{Z}(v) \neq \emptyset\), the selected subset is:
\begin{equation}
    S_v = \operatorname{top}_{B}\!\Big(\mathcal{Z}(v),\;
    \mathrm{KL}\big(\tilde{q}_j \,\|\, p\big)\Big).
\end{equation}
The final variable set is \(\bigcup_{v \in \mathcal{V}_L} S_v\).

\section{Limitations}
\label{app:limitations}

\subsection{Variable Proposal and Captioning}

The quality of proposed variables and captions is bounded by the VLM's pretraining corpus. Domains or taxa that are poorly represented in training data may yield superficial or inaccurate variables, and the model's familiarity is not uniform across the label space. Well-studied organisms (e.g., hard corals) are likely to be better detected than understudied groups (e.g., soft corals), introducing a systematic bias in caption informativeness. 

The uniform per-node budget \(B\) and conservative variable selection reflect pragmatic constraints imposed by the limited token capacity of smaller VLMs. When provided with too many variables, models at this scale tend to anchor onto variable names without meaningful contextual processing \citep{llms_easily_distracted, llms_do_in-context_learning_differently}. A larger VLM or a multi-pass captioning scheme could improve caption quality. Lastly, while it may only need to be performed once, the captioning cost may be quite high, as demonstrated in \autoref{tab:task_diagnostics}. Computation cost in this step can also vary significantly depending on the phrasing used in the prompt.

\subsection{Architecture and Training}

The text and image embedding models (Qwen3-Embedding-8B and DINOv3 ViT-H respectively) are frozen throughout training. Their representation spaces are therefore not adapted to the target domain, and any distributional mismatch between pretraining data and benthic imagery is inherited by the downstream classifier. Fine-tuning these backbones could improve performance but would substantially increase computational requirements and reduce accessibility for domain-specialised labs with limited hardware.

The router models introduce additional parameters relative to the unimodal baselines (by a factor of 3, given each stream, and the router mechanism), and one variant (Router) loads pretrained pooling layers from unimodal runs. We therefore assess the text contribution with ablations that hold the architecture and parameter count fixed while removing the per-sample text signal (\autoref{subsec:subtree_prior}; \autoref{app:noise}).

\subsection{Evaluation Scope}

Our benthic experiments, our primary application focus, are restricted to two datasets (BenthicNet \citep{benthicnet} and FathomNet (\citealp{fathomnet}; \citetalias{fathomnet-2025})), which represent the few large-scale, globally diverse, and taxonomically broad benchmarks available in this field. Beyond the benthic domain, we evaluate only on iNaturalist. While the core methodology does not rely on domain-specific assumptions, its extension to other specialised domains, such as medical imaging, remains empirically undemonstrated and is left to future work.

\subsection{Evaluation Metric}

The inverted HML score \(\dinv\) relies on a user-specified parameter \(k_\text{P}\) that governs the severity of the cardinality penalty. Our choice of \(k_\text{P} = 0.5\) represents a balanced default. Where the expected unannotated multi-label cardinality of a dataset is suspected, \(k_\text{P}\) can be adjusted to reflect it.

Additionally, \(\dinv\) treats all edges in the hierarchy as having unit weight, which may not reflect the true semantic distance between nodes. For example, the distinction between two congeneric species is typically far smaller than that between two phyla, yet both correspond to two edge traversals in a taxonomic tree. Depth-specific edge weights could be incorporated where expert knowledge provides meaningful inter-node distances, but we do not explore this idea in this work.

\section{Additional Ablations}
\label{app:additional_ablations}
 
\subsection{Zero-Shot VLM Classification}
\label{app:zeroshot}
\input{tables/table8_zero_shot_vlm}
 
A natural question is whether the VLM can perform the classification directly, without the elicitation and training pipeline. We evaluate this on BenthicNet in two prompting regimes under greedy decoding, and report structural information rather than performance (\autoref{tab:zero_shot_vlm}).
 
Free-form output cannot be scored. Both macro \(F_1\) and \(\dinv\) require an ancestor-closed subset of \(\mathcal{V}\), whereas the VLM emits unconstrained text. Matching that text back to the scheme recovers 0.10 nodes per image on BN-Sub and 0.87 on BN-Bio, against ground truth cardinalities of 1.42 and 3.09. Only 0.5\% and 3.3\% of emitted terms correspond to any CATAMI category, and 92.0\% and 77.6\% of images yield no scheme-conformant prediction at all. Any score reported under this regime would be a property of the matching rule rather than of the model.
 
Supplying the vocabulary in the prompt helps with conformance but worsens cardinality. Conformance rises to 99.9\% on BN-Sub and 98.0\% on BN-Bio, while predicted cardinality inflates to 2.08 and 6.98 times ground truth. On BN-Bio the failure is bimodal: the median image returns 6 categories, 6.3\% return more than 54, and the worst case returns 148 against a single annotated endpoint. This is the regime \(\dinv\) penalises and micro-averaged AP tolerates (\autoref{app:hml_metrics_study}): conformant predictions that share correct ancestors while carrying almost no discriminative information. It is also the reason a per-node budget \(B\) is necessary (\autoref{app:latent_extraction_details}), since instability scales with the size of the vocabulary placed in the prompt.
 
Neither regime yields a usable classifier. Knowledge held in the weights does not make a specialised label space available as a prediction target, which motivates using the VLM to elicit properties rather than to predict nodes.
 
\subsection{Scale-Matched Noise Ablation}
\label{app:noise}
\input{tables/table10_noise_ablation}
 
To separate the contribution of the text content from that of the additional input dimensions, we replace the text embedding with Gaussian noise of matched scale at a mixing coefficient of one, so that no original signal remains, and retrain under the identical Router protocol. Architecture, embedding dimensionality and trainable parameter count are held constant, and the model trains from the start on an uninformative stream, free to disregard it in favour of the image branch.
 
Results are given in \autoref{tab:noise_ablation}. On both BenthicNet domains and on iNat the corrupted model degrades on both metrics at every seed, and falls below the image-only baseline of \autoref{tab:results} and \autoref{tab:inat_results}. Additional feature dimensions should not make a model worse than not having them, so a gain arising from dimensionality alone would not produce this. This ablation separates content from dimensionality. A comparison against other semantically valid captions is given in \autoref{app:caption_probe}.
 
On FathomNet the effect is significant but small, and macro \(F_1\) is unchanged. Two factors plausibly limit what the VLM can supply for this dataset: annotations given as Latin binomials rather than descriptive terms, and the limited representation of deep-sea organisms in general-purpose corpora relative to terrestrial fauna. We do not attempt to separate these.
 
\subsection{Descriptor-Matching Baseline}
\label{app:dclip}
\input{tables/table7_dclip}

We compare against a descriptor-matching baseline following \citet{dclip}, which classifies images by their similarity to LLM-generated visual descriptors rather than to class names alone. Because the method produces flat outputs, we evaluate it both with and without the hierarchical constraint of \autoref{sec:methods}.

For each of the 293 BN-Bio nodes, the VLM used elsewhere in our pipeline generates six visual descriptors, prompted with the node and its position in the hierarchy and restricted to appearance, giving 1,758 descriptors in total. Each descriptor is embedded once with the CLIP text encoder and scored against every image by cosine similarity, and a node's score is the mean over its six descriptors. Normalisation and per-node thresholds are fitted on the 105,133 training images and applied unchanged to the 37,504 test images, so the baseline receives the same per-node thresholding as every model in \autoref{tab:results}.

The baseline reaches roughly half the macro \(F_1\) of the trained image branch and under a tenth of its \(\dinv\) (\autoref{tab:dclip}). Adding the hierarchical constraint leaves macro \(F_1\) unchanged and raises macro \(\dinv\) from 0.12 to 0.17, while the sample-mean \(\dinv\) stays at 0.11, so the gain falls on nodes that carry little weight in the sample mean.

The descriptors account for much of this. Of the 1,758 generated, 22.5\% are verbatim duplicates of a descriptor issued for another node: \textit{smooth or slightly rough surface} is assigned to 35 nodes and \textit{pale beige to brownish hue} to 47. Duplication is most common between siblings, where 24\% of the 769 sibling pairs share an identical descriptor. For example, \textit{Macroalgae \(>\) Laminate \(>\) Brown} and \textit{\(>\) Red} receive five identical descriptors out of six, none of which mentions colour, although the prompt supplied both hierarchy paths and asked for distinguishing features. Node scores reflect this overlap: sibling nodes correlate at \(r = 0.825\), against \(0.469\) for non-siblings, and the correlation among nodes at the same depth rises from \(0.36\) at depth one to \(0.95\) at depth six. In this domain, appearance-based descriptors separate coarse categories but not the morphological distinctions on which CATAMI subdivides.

Our variables also operate at the clade level; the difference lies in redundancy. A visual descriptor names something a frozen visual encoder already extracts, whereas the properties elicited in \autoref{sec:methods} are not recoverable from the image alone. This is consistent with \autoref{app:caption_probe}, where, at the coarsest level of the hierarchy, generic image descriptions are markedly more redundant with the image embedding than captions built on the elicited variables.

\section{BN-Bio (OOD) Case Study}
\label{app:bn-bio_case_study}
\input{figures/fig4_text_tow/fig4_text_tow}
\input{figures/fig4-5_property_correlations/fig4-5_property_correlations}

The correlation analysis between proposed variables and text advantage (\autoref{fig:property_correlation_bars}) reveals that behavioural and habitat traits (\textit{ambush}, \textit{high mobility}, \textit{benthic habitat}) are significantly positively correlated with text modality performance gains, while morphological plant traits (\textit{light interception}, \textit{leaf morphology}) exhibit significant negative correlations. However, examining text advantage at the level of individual nodes (\autoref{fig:property_class_pairs}) reveals that a variable annotated as absent (Property: [0]) for a given sample may still contribute positively to the discriminability of relevant classes. The explicit negation of a property, along with its caption, may serve as part of a combinatorial signature that distinguishes a node from its siblings.

Despite this argument, we emphasise that these correlations are observational and do not establish a causal mechanism by which any individual variable improves classification. While connections can be manually verified in specific cases, we cannot generalise this reasoning to all variable–node pairs. Some variables may contribute indirectly through interactions with other annotations, or may serve primarily as prompt structure that encourages the captioning VLM to attend to relevant image regions. Disentangling these effects would require controlled ablations at the individual variable level, beyond the scope of this work.

\subsection{Node-Wise Analysis}
\label{app:failure}

To characterise where the text modality contributes, we compare the text and image branches node by node on every BN-Bio node with at least 20 positive test samples (171 nodes).

On its own, the text branch has lower \(F_1\) than the image branch at 157 of the 171 nodes. The gap is recall-driven at 105 of these nodes and precision-driven at 52, so captions more often omit a discriminating property than assert an incorrect one. This matches the clade-level character of the elicited variables: \textit{filter feeding}, for example, separates sponges from other taxa, placing a prediction within the correct subtree without resolving the morphotype within it.

The two metrics diverge in a consistent direction. At 75 of the 171 nodes, the text branch lands closer to the target on \(\dinv\) while failing to identify the node on \(F_1\); the reverse holds at only 6. Text therefore contributes hierarchical proximity where it does not contribute discrimination.

In the fused model, routing recovers the text branch's \(F_1\) deficit at 130 of the 157 nodes, and at all 21 measurable BN-Sub nodes, with a median gain of 0.018 \(F_1\) over the image branch. At the remaining 27 nodes, the median cost is 0.005 \(F_1\). These nodes are concentrated in sponges, which make up 24.6\% of the nodes examined but 55.6\% of those not recovered, a 2.3-fold concentration. CATAMI subdivides sponges by morphology, which our proposal prompt excludes by design (\autoref{app:prompt_proposal}), so these nodes fall where the clade-level character of the variables predicts.
 
\subsection{Gate Weights}
\label{app:gate}
\input{tables/table9_gate_weights}

The router assigns per-node weights over the image, text and cross-attention branches. \autoref{tab:gate} reports the mean text weight and its per-sample standard deviation with and without per-sample text information.

The gate does not ignore the text branch. On BN-Bio, the mean text weight is \(0.57\). When every caption is replaced by the same payload, the mean text weight falls by \(3.6\%\) (\(0.566 \rightarrow 0.546\)), lower at every seed, while its per-sample standard deviation changes by less than \(1\%\). The sample-to-sample variation of the gate is therefore driven mainly by the image side of its input rather than by the content of individual captions.

These weights describe how much the fused output draws on each branch, not how the text branch transforms its input. Because the gate mixes branch logits, it does not show how the text embedding is used within the model; that contribution is measured directly by the ablations in \autoref{subsec:subtree_prior} and \autoref{app:noise}.

\section{HML Metrics Noise Experiment}
\label{app:hml_metrics_study}
\input{algorithms/algorithm2a_random_hierarchy_generation}
\input{algorithms/algorithm2b_random_hierarchical_data_generation}
\input{algorithms/algorithm2c_realistic_hml_noise_generation}
\input{figures/fig5_metrics_comparison/fig5a_noise_effect}
\input{figures/fig5_metrics_comparison/fig5b_metrics_results}

A natural notion of a ``random'' model in HML is one that predicts each node independently with uniform probability, analogous to the random baseline in multi-class classification. Such a model produces predictions with expected cardinality near \(|\mathcal{V}|/2\), where the number of combinatorial states is greatest. Crucially, the hierarchical constraint forces false positives to propagate upward since activating a node requires activating all of its ancestors. Consequently, higher-level nodes are disproportionately likely to be active. This stacking effect means that even moderate noise inflates cardinality rapidly. Both AP and \(F_1\) decay under this noise regime, and do so more steeply as hierarchy size grows, since the propagation affects a larger number of ancestor nodes in absolute terms. Recall, however, remains high throughout, because the noise mechanism adds false positives without removing true positives. Instead, the decrease in \(F_1\) is driven mostly by precision. By contrast, \(\dinv\) falls off sharply under this noise, owing to its direct penalisation of excessive prediction cardinality, correctly reflecting that a model whose cardinality has inflated toward \(|\mathcal{V}|/2\) retains little predictive value. The key point is not that one metric is right and the other wrong here, but that this regime is the \emph{only} one in which AP behaves acceptably.

Indeed, this type of performance degradation is almost never relevant in real-world scenarios. HML classification is a difficult problem because natural imbalances arise at deeper nodes \citep{improving_detection_rare_nodes, hml_imbalance_hros}, with detection and recall falling in these specific and rare regions. In contrast, higher-level nodes, being more frequently observed, are rarely confused with one another. Instead, a realistic noise environment sharply increases the rate of false negatives with node depth, and only slightly increases false positives, which decrease with depth. Following these guidelines, we simulate data by generating synthetic hierarchies as in \autoref{alg:hierarchy_gen}, with fictitious data generated using \autoref{alg:data_gen}, and add noise with \autoref{alg:noise_gen}. The effects of the added noise are shown in \autoref{fig:noise_experiment_profile}. Our synthetic experiment targets an expected cardinality of three (to simulate the low cardinality and sparsity of benthic data). We tested five distinct hierarchies at each of five sizes (25 hierarchies in total), with 500 data points per hierarchy. The noise levels spanned \(0.0\) to \(1.0\) in \(0.1\) increments.

In this more realistic setting (\autoref{fig:noise_experiment_metrics_comparison}), we observe that AP is both insensitive to disruptive noise and misleading as to model quality, leading to an optimistic overestimation of model performance. This behaviour matches results in existing studies \citep{hml_missing_info_benthicnet, coherent_hml, hml_imbalance_hros, improving_detection_rare_nodes}. AP behaves this way because of the sparsity of nodes in large hierarchies. Low ground-truth cardinality means that the vast majority of nodes are inactive for any given sample. Micro-average AP computes precision and recall across all node-sample pairs simultaneously, so by its inability to distinguish the trivial discrimination of inactive nodes from the genuinely difficult and important task of detecting active ones, the metric becomes inflated. On the other hand, \(\dinv\) is not susceptible to this inflation because it operates directly on predicted and ground-truth cardinalities. Consequently, \(\dinv\) degrades more steeply than both \(F_1\) and AP as noise intensifies, reflecting the practical reality that the model has lost its ability to resolve the fine-grained distinctions important to HML classification.

\subsection{Comparisons to Hierarchical \texorpdfstring{\(F_1\)}{F1}}
\label{app:sub:inv_hml_dist_vs_hf1}

\input{figures/fig6_hf1_d_inv_hierarchy/fig6_hf1_d_inv_hierarchy}

We compare \(\dinv\) and \(hF_1\) \citep{hml_econ_lit} on two possible prediction strategies that we dub ``sibling hedging'' and ``depth over-prediction''. Throughout this section, we refer to the example hierarchy depicted in \autoref{fig:hf1_comparison}. Consider a hierarchy with root node \(r\), internal nodes \(A\) and \(B\) as children of \(r\), and leaf nodes \(a_1, a_2, a_3\) as children of \(A\) and \(b_1, b_2\) as children of \(B\). The hierarchy therefore contains \(|\mathcal{V}|=8\) nodes. Suppose the ground truth annotation for a given sample is \(\mathcal{Y} = \{a_1\}\), so \(n_{\text{truth}} = 1\). The maximum tree distance from \(a_1\) to any other node is \(d_{\max} = 4\), attainable by \(b_1\) and \(b_2\). For reference, the perfect prediction \(\hat{\mathcal{Y}} = \{a_1\}\) achieves \(hF_1 = 1.0\) and \(\dinv = 1.0\).

First, consider sibling hedging, where a risk-averse classifier predicts all siblings of the true label, producing \(\hat{\mathcal{Y}} = \{a_1, a_2, a_3\}\) with \(n_{\text{pred}} = 3\). To compute \(hF_1\), each label is expanded into the set of itself and its ancestors. The expanded ground truth is \(\mathcal{Y}^{\uparrow} = \{a_1, A\}\) (ignoring the root) and the expanded prediction is \(\hat{\mathcal{Y}}^{\uparrow} = \{a_1, a_2, a_3, A\}\). Precision is \(|\mathcal{Y}^{\uparrow} \cap \hat{\mathcal{Y}}^{\uparrow}| \;/\; |\hat{\mathcal{Y}}^{\uparrow}| = |\{a_1, A\}| \;/\; |\{a_1, a_2, a_3, A\}| = 2/4\), and recall is \(|\mathcal{Y}^{\uparrow} \cap \hat{\mathcal{Y}}^{\uparrow}| \;/\; |\mathcal{Y}^{\uparrow}| = 2/2 = 1\). The harmonic mean gives \(hF_1 = 2 \cdot (1/2) \cdot 1 \;/\; (1/2 + 1) = 1 \;/\; (3/2) = 2/3 = 0.67\). The ancestor overlap substantially attenuates the penalty for the two spurious predictions, because siblings share their full ancestral lineage with the true label. This leniency intensifies as hierarchy depth increases relative to the count of sibling nodes, since the shared ancestral path increasingly dominates the expanded sets.

To compute \(\dinv\) for the same scenario, we first construct the score matrix \(\mathbf{D}_{\text{inv}} \in \mathbb{R}^{n_{\text{truth}} \times n_{\text{pred}}} = \mathbb{R}^{1 \times 3}\), where each entry is the normalised similarity \((d_{\max} - d_{\text{pred}}) / d_{\max}\). The tree distances from \(a_1\) to each prediction are \(d_{\mathcal{T}}(a_1, a_1) = 0\), \(d_{\mathcal{T}}(a_1, a_2) = 2\), and \(d_{\mathcal{T}}(a_1, a_3) = 2\), giving \(\mathbf{D}_{\text{inv}} = \begin{bmatrix} (4{-}0)/4 & (4{-}2)/4 & (4{-}2)/4 \end{bmatrix} = \begin{bmatrix} 1.0 & 0.5 & 0.5 \end{bmatrix}\). Solving the rectangular assignment problem yields \(f_{\text{JV}}(\hat{\mathcal{Y}}, \mathcal{Y};\, \mathbf{D}_{\text{inv}}) = 1.0\), so \(s_{\max} = f_{\text{JV}}/n_{\text{truth}} = 1.0\), since the optimal match assigns predicted \(a_1\) to true \(a_1\) at a score of \(1.0\). For the cardinality factor, we have \(n_{\text{E}} = n_{\text{pred}} - n_{\text{truth}} = 3 - 1 = 2\) excess predictions and \(n_{\text{ME}} = |\mathcal{V}| - n_{\text{truth}} = 8 - 1 = 7\). Evaluating \(\gamma(2;\, k_{\text{P}}{=}0.5,\, n_{\text{ME}}{=}7)\) numerically via \autoref{eq:exponential_decay_form} gives \(\gamma \approx 0.248\). The final score is therefore \(\dinv = s_{\max} \cdot \gamma = 1.0 \cdot 0.248 \approx 0.25\). While \(hF_1\) awards \(0.67\) to this prediction, \(\dinv\) assigns only \(0.25\), reflecting the intuition that predicting all children when only one is correct constitutes a significant cardinality violation.

Now consider depth over-prediction, where the classifier predicts every leaf: \(\hat{\mathcal{Y}} = \{a_1, a_2, a_3, b_1, b_2\}\) with \(n_{\text{pred}} = 5\). For \(hF_1\), the expanded prediction becomes \(\hat{\mathcal{Y}}^{\uparrow} = \{a_1, a_2, a_3, b_1, b_2, A, B\}\), containing seven elements, while the expanded ground truth remains \(\mathcal{Y}^{\uparrow} = \{a_1, A\}\). Precision is \(|\{a_1, A\}| \;/\; |\{a_1, a_2, a_3, b_1, b_2, A, B\}| = 2/7\), recall is \(2/2 = 1\), and \(hF_1 = 2 \cdot (2/7) \cdot 1 \;/\; (2/7 + 1) = (4/7) \;/\; (9/7) = 4/9 \approx 0.44\). Despite predicting every leaf, the metric remains well above zero because several nodes share intermediate ancestors.

For \(\dinv\), the score matrix is \(\mathbf{D}_{\text{inv}} \in \mathbb{R}^{1 \times 5}\). The tree distances from \(a_1\) are \(d_{\mathcal{T}}(a_1, a_1) = 0\), \(d_{\mathcal{T}}(a_1, a_2) = 2\), \(d_{\mathcal{T}}(a_1, a_3) = 2\), \(d_{\mathcal{T}}(a_1, b_1) = 4\), and \(d_{\mathcal{T}}(a_1, b_2) = 4\), yielding \(\mathbf{D}_{\text{inv}} = \begin{bmatrix} 1.0 & 0.5 & 0.5 & 0.0 & 0.0 \end{bmatrix}\). The optimal assignment again matches predicted \(a_1\) to true \(a_1\), giving \(f_{\text{JV}} = 1.0\) and hence \(s_{\max} = f_{\text{JV}}/n_{\text{truth}} = 1.0\). Now \(n_{\text{E}} = 5 - 1 = 4\) and \(n_{\text{ME}} = 7\), so the cardinality factor is \(\gamma(4;\, 0.5,\, 7) \approx 0.057\), and the final score is \(\dinv = 1.0 \cdot 0.057 \approx 0.06\). Where \(hF_1\) still awards \(0.44\) to a degenerate prediction that labels every leaf in the hierarchy, \(\dinv\) reduces the score to near zero, faithfully capturing that such a prediction carries almost no discriminative information.

The divergence between the two metrics in these scenarios stems from a fundamental design difference. Because \(hF_1\) operates on ancestor-expanded sets without solving an explicit assignment between predicted and true labels, any overlap in ancestral paths contributes positively to both precision and recall, regardless of whether those shared ancestors reflect genuine predictive accuracy or are simply artefacts of the hierarchical structure. In shallow hierarchies with few shared ancestors this effect can be minor, but in deep or broad hierarchies it can substantially dampen the penalty for spurious predictions. By contrast, \(\dinv\) isolates the best possible matching through the assignment problem and then applies the cardinality factor \(\gamma\), which decays monotonically with the number of excess predictions. This ensures that cardinality violations are penalised in proportion to their severity, independently of where in the hierarchy the spurious predictions happen to fall. We view this property as particularly important for applications where downstream decisions depend on the precise set of predicted labels rather than merely on their neighbourhood in the hierarchy.

This leniency has been identified independently. \citet{cophe} observe the same structural property in set-based hierarchical evaluation, of which \(hF_1\) is an instance. Because the expanded sets are unions, several predicted descendants of a common ancestor reduce to a single entry, and the measure does not consider how many predictions reach the ancestor. They address this by propagating descendant counts rather than truth values, and report that the resulting measure returns consistently lower scores than set-based evaluation across three trained models on the MIMIC-III ICD-9 task, attributing the gap to over- and under-prediction within families of codes that set-based evaluation cannot express.

Our own results are in agreement. In \autoref{tab:decomposition}, the degradation observed under text ablation is registered most weakly by \(hF_1\) among the cardinality-aware scores, and disappears entirely once \(\gamma\) is fixed to one, indicating that the change being measured is one of prediction cardinality. That two measures constructed for different domains exhibit the same insensitivity suggests it follows from ancestor expansion itself rather than from any particular hierarchy or dataset.

\subsection{Decomposing the \texorpdfstring{\(\dinv\)}{d\_inv} Ablation}
\label{app:decomposition}
\input{tables/table6-5_decomposition}

Removing the per-sample text signal lowers \(\dinv\) on BN-Bio by 21.7\% (\autoref{tab:mean_replacement}), but this alone does not show which component of \(\dinv\) is responsible. \(\dinv\) is the product of a proximity term \(s_{\max}\), the assignment-matched hierarchical similarity, and a cardinality factor \(\gamma\), which penalises excess predictions. We therefore rescore the BN-Bio mean-replacement runs under alternative cardinality factors, holding the matched assignment fixed so that only the cardinality term changes (\autoref{tab:decomposition}).

The degradation appears under every cardinality-aware score, including three that we did not introduce: a min/max ratio, an \(F_1\)-style count penalty, and \(hF_1\), which registers it most weakly. Setting \(\gamma \equiv 1\), which leaves proximity alone, removes the degradation, while the mean number of predicted endpoints rises from 10.16 to 10.29 against a ground-truth mean of 2.34. The change is therefore attributable to excess prediction cardinality rather than to matched proximity. This is the signature expected of a sub-tree prior. Since \(s_{\max}\) scores only the best-matched prediction for each ground-truth endpoint, suppressing predictions in incorrect subtrees can only come through in \(\gamma\). This concerns the fused model with and without per-sample text, and is distinct from the proximity advantage in \autoref{subsec:results_and_discussion}, which compares the text branch with the image branch.

The weak response of \(hF_1\) matches \autoref{app:sub:inv_hml_dist_vs_hf1}. Because \(hF_1\) expands each label to itself and its ancestors, spurious predictions that share a lineage with the target are only lightly penalised, so a change expressed through cardinality is precisely the kind to which \(hF_1\) is least sensitive.

\section{Additional Details for Inverted HML Score}
\label{app:inv_hml_dist_proofs}

This section contains edge case definitions and proofs for the main claims made in the introduction of the inverted HML score (\(\dinv\)) in \autoref{sec:methods}. These claims and proofs serve to demonstrate the mathematical behaviour and robustness of the cardinality component of \(\dinv\).

\subsection{Edge Cases}

Three edge cases for \(\dinv\) are defined as follows. Where no path exists between a ground-truth endpoint and a predicted one, as is the case between disjoint roots, the pairwise similarity is set to zero rather than left undefined. Where the prediction set is empty, \(\dinv = 0\). Where the prediction is deficient rather than excessive, \(n_\text{E} \leq 0\) and \(\gamma = 1\), so under-prediction is penalised instead in \(s_{\max}\) through unmatched ground-truth components, rather than in the cardinality factor.
\newline
\input{theorems/thm1_proof_of_kp_star}
\input{theorems/thm2_proof_of_lin_for_kp_star}
\input{theorems/thm3_proof_of_b_exist}


\end{document}

%% file: math_commands.tex
\usepackage{amsmath,amsfonts,bm}

\def\eqref#1{equation~\ref{#1}}

\def\1{\bm{1}}

\DeclareMathAlphabet{\mathsfit}{\encodingdefault}{\sfdefault}{m}{sl}
\SetMathAlphabet{\mathsfit}{bold}{\encodingdefault}{\sfdefault}{bx}{n}



%% file: figures/fig1_ghiml_motivation/fig1_ghiml_motivation.tex
\begin{figure}[!htbp]
    \centering
    \includegraphics[width=0.8\linewidth]{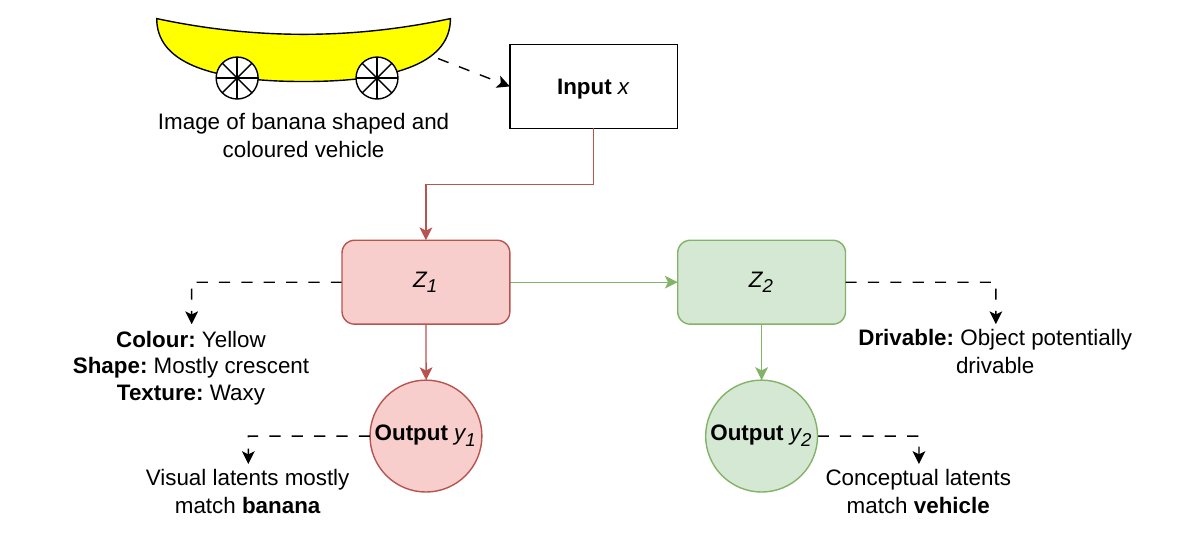}
    \caption{\textbf{Difference in Latents.} A hypothetical example. The red (left) branch depicts classification typical of contemporary vision-based neural networks, whose low-level latents learned from a limited dataset are insufficient to classify the object correctly. The green (right) branch additionally draws on general conceptual knowledge that cannot be learned from visual characteristics alone.}
    \label{fig:ghiml_motivation}
\end{figure}

%% file: figures/fig2_ghiml_captioning/fig2_ghiml_captioning.tex
\begin{figure}[!htbp]
    \centering
    \includegraphics[width=0.7\linewidth]{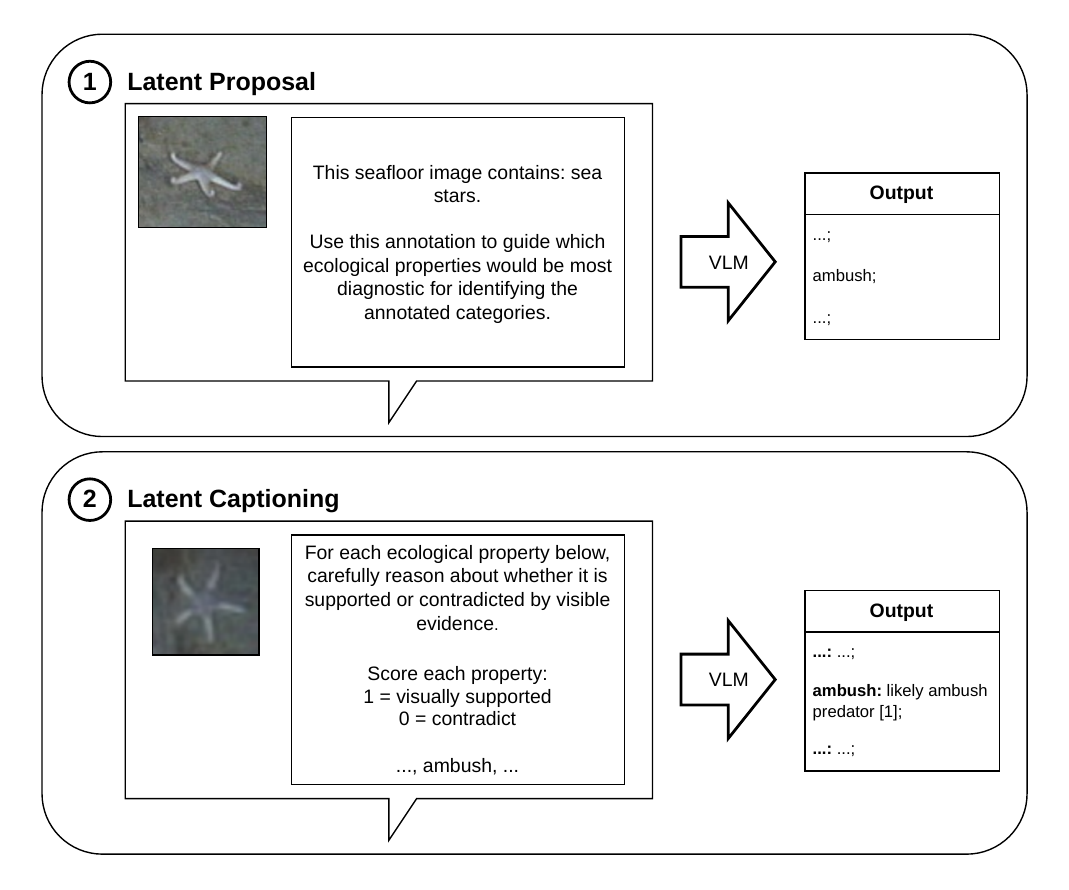}
    \caption{\textbf{VLM Variable Proposal and Captioning.} A sea star image and its annotation are used to propose the variable ``ambush''; during captioning, the VLM is prompted to reason about each variable for each sample.}
    \label{fig:ghiml_captioning}
\end{figure}

%% file: figures/fig3_architectures/fig3_architectures.tex
\begin{figure}[!htbp]
    \centering
    \begin{minipage}[c]{0.52\linewidth}
        \centering
        \includegraphics[width=\linewidth]{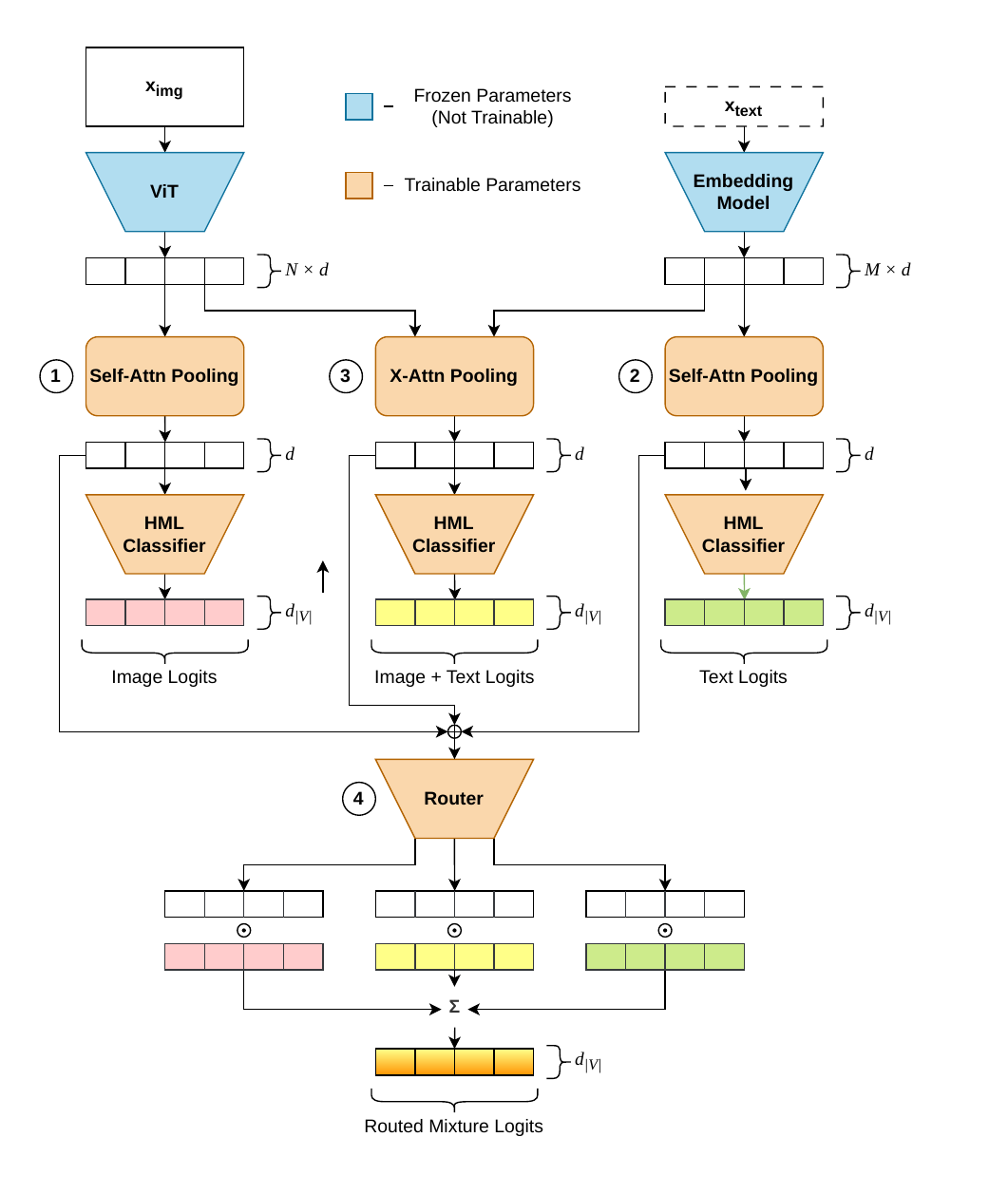}
    \end{minipage}\hfill
    \begin{minipage}[c]{0.44\linewidth}
        \caption{\textbf{Model Architectures.} The (1) image and (2) text streams apply self-attention pooling followed by an HML-constrained classifier; (3) applies cross-attention followed by pooling; (4) a router predicts per-node weights that mix the logits of (1)--(3).}
        \label{fig:architecture}
    \end{minipage}
\end{figure}

%% file: tables/table1_benthicnet.tex
\begin{table}[t]
    \centering
    \caption{\textbf{HML Results.} Macro-average performance across datasets, modalities, and distribution settings.
    Values are mean \(\pm\) std (\%).
    Superscript indicates mean rank across seeds.
    Ten seeds were tested for BN datasets, and thirty for FN.
    Coloured values differ significantly from Img alone (Wilcoxon signed-rank test;
    \sigA{green}: \(p < 0.05\);
    \sigB{bold dark green}: \(p < 0.01\)).}
    \label{tab:results}
    \resizebox{\textwidth}{!}{%
    \begin{tabular}{l l r@{}l r@{}l r@{}l r@{}l r@{}l r@{}l}
        \toprule
        & 
            & \multicolumn{4}{c}{\textbf{BN-Sub}} 
            & \multicolumn{4}{c}{\textbf{BN-Bio}}
            & \multicolumn{4}{c}{\textbf{FN}} \\
        \cmidrule(lr){3-6}
        \cmidrule(lr){7-10}
        \cmidrule(lr){11-14}
        & 
            & \multicolumn{2}{c}{\textbf{F1}} 
            & \multicolumn{2}{c}{\textbf{\(d_\text{inv}\)}} 
            & \multicolumn{2}{c}{\textbf{F1}} 
            & \multicolumn{2}{c}{\textbf{\(d_\text{inv}\)}}
            & \multicolumn{2}{c}{\textbf{F1}} 
            & \multicolumn{2}{c}{\textbf{\(d_\text{inv}\)}} \\
        \midrule
        \multirow{5}{*}{\rotatebox{90}{\textbf{ID}}}
        & Txt 
            & \(48.5_{\pm 2.9}\) & \textsuperscript{(5.0)} 
            & \(40.5_{\pm 5.3}\) & \textsuperscript{(5.0)} 
            & \(7.8_{\pm 1.6}\) & \textsuperscript{(5.0)} 
            & \(1.2_{\pm 1.7}\) & \textsuperscript{(5.0)}
            & \(15.6_{\pm 2.9}\) & \textsuperscript{(5.0)} 
            & \(0.3_{\pm 0.2}\) & \textsuperscript{(5.0)} \\
        & Img 
            & \(77.9_{\pm 3.5}\) & \textsuperscript{(3.4)} 
            & \(68.1_{\pm 4.5}\) & \textsuperscript{(3.5)} 
            & \(14.3_{\pm 3.8}\) & \textsuperscript{(3.2)} 
            & \(5.1_{\pm 5.5}\) & \textsuperscript{(3.5)}
            & \(44.5_{\pm 5.1}\) & \textsuperscript{(3.5)} 
            & \(16.5_{\pm 3.2}\) & \textsuperscript{(3.5)} \\
        & Img + Txt 
            & \(78.3_{\pm 3.7}\) & \textsuperscript{(3.0)} 
            & \(68.7_{\pm 4.3}\) & \textsuperscript{(3.0)} 
            & \(14.0_{\pm 3.5}\) & \textsuperscript{(3.8)} 
            & \(4.9_{\pm 5.6}\) & \textsuperscript{(3.5)}
            & \(44.5_{\pm 5.2}\) & \textsuperscript{(3.5)} 
            & \(16.7_{\pm 3.3}\) & \textsuperscript{(3.4)} \\
        & Router 
            & \sigB{\(81.4_{\pm 2.5}\)} & \sigB{\textsuperscript{(1.1)}} 
            & \sigB{\(72.7_{\pm 3.0}\)} & \sigB{\textsuperscript{(1.1)}} 
            & \sigB{\(16.6_{\pm 3.5}\)} & \sigB{\textsuperscript{(2.0)}} 
            & \sigB{\(7.6_{\pm 6.5}\)} & \sigB{\textsuperscript{(1.5)}}
            & \sigB{\(51.2_{\pm 4.2}\)} & \sigB{\textsuperscript{(1.1)}} 
            & \sigB{\(37.6_{\pm 4.1}\)} & \sigB{\textsuperscript{(1.0)}} \\
        & Router\textsuperscript{\dag} 
            & \sigA{\(79.4_{\pm 2.3}\)} & \sigA{\textsuperscript{(2.5)}} 
            & \sigA{\(69.4_{\pm 3.9}\)} & \sigA{\textsuperscript{(2.4)}} 
            & \sigB{\(17.8_{\pm 4.0}\)} & \sigB{\textsuperscript{(1.0)}} 
            & \sigB{\(7.7_{\pm 6.9}\)} & \sigB{\textsuperscript{(1.5)}}
            & \sigB{\(49.7_{\pm 4.6}\)} & \sigB{\textsuperscript{(1.9)}} 
            & \sigB{\(23.0_{\pm 4.6}\)} & \sigB{\textsuperscript{(2.0)}} \\
        \midrule
        \multirow{5}{*}{\rotatebox{90}{\textbf{OOD}}}
        & Txt 
            & \(38.6_{\pm 3.2}\) & \textsuperscript{(5.0)} 
            & \(36.6_{\pm 5.7}\) & \textsuperscript{(5.0)} 
            & \(7.4_{\pm 1.5}\) & \textsuperscript{(5.0)} 
            & \(2.5_{\pm 3.4}\) & \textsuperscript{(4.8)}
            & \(12.4_{\pm 4.8}\) & \textsuperscript{(5.0)} 
            & \(0.4_{\pm 0.3}\) & \textsuperscript{(5.0)} \\
        & Img
            & \(64.8_{\pm 4.5}\) & \textsuperscript{(2.5)} 
            & \(60.2_{\pm 5.3}\) & \textsuperscript{(3.2)} 
            & \(11.4_{\pm 2.7}\) & \textsuperscript{(3.4)} 
            & \(4.4_{\pm 5.1}\) & \textsuperscript{(3.8)}
            & \(37.4_{\pm 10.0}\) & \textsuperscript{(3.4)} 
            & \(18.3_{\pm 4.8}\) & \textsuperscript{(3.5)} \\
        & Img + Txt 
            & \(64.2_{\pm 4.6}\) & \textsuperscript{(3.1)} 
            & \(60.4_{\pm 5.1}\) & \textsuperscript{(2.7)} 
            & \(11.3_{\pm 2.8}\) & \textsuperscript{(3.6)} 
            & \sigA{\(4.7_{\pm 5.8}\)} & \sigA{\textsuperscript{(3.0)}}
            & \(37.4_{\pm 10.0}\) & \textsuperscript{(3.4)} 
            & \(18.4_{\pm 5.1}\) & \textsuperscript{(3.3)} \\
        & Router 
            & \sigB{\(67.7_{\pm 3.1}\)} & \sigB{\textsuperscript{(1.1)}} 
            & \sigB{\(64.5_{\pm 3.6}\)} & \sigB{\textsuperscript{(1.0)}} 
            & \sigB{\(13.2_{\pm 3.0}\)} & \sigB{\textsuperscript{(1.9)}} 
            & \sigB{\(5.9_{\pm 5.8}\)} & \sigB{\textsuperscript{(2.0)}}
            & \sigB{\(40.9_{\pm 9.2}\)} & \sigB{\textsuperscript{(1.7)}} 
            & \sigB{\(32.5_{\pm 6.4}\)} & \sigB{\textsuperscript{(1.0)}} \\
        & Router\textsuperscript{\dag} 
            & \(64.0_{\pm 4.8}\) & \textsuperscript{(3.3)} 
            & \(60.4_{\pm 5.3}\) & \textsuperscript{(3.1)} 
            & \sigB{\(14.3_{\pm 3.7}\)} & \sigB{\textsuperscript{(1.1)}} 
            & \sigB{\(6.4_{\pm 6.4}\)} & \sigB{\textsuperscript{(1.4)}}
            & \sigB{\(41.7_{\pm 9.2}\)} & \sigB{\textsuperscript{(1.5)}} 
            & \sigB{\(23.1_{\pm 6.5}\)} & \sigB{\textsuperscript{(2.2)}} \\
        \bottomrule
    \end{tabular}%
    }
\end{table}

%% file: tables/table1-5_inat.tex
\begin{table}[!htbp]
    \centering
    \caption{\textbf{iNaturalist Results.} Macro-average performance over the 100,000 held-out images. Values are mean \(\pm\) std (\%) across five seeds; superscript indicates mean rank across seeds. Coloured values are significantly higher than Img (one-sided Wilcoxon signed-rank test; \sigA{green}: \(p < 0.05\)).}
    \label{tab:inat_results}
    \begin{tabular}{l r@{}l r@{}l}
        \toprule
        & \multicolumn{2}{c}{\textbf{F1}}
        & \multicolumn{2}{c}{\textbf{\(d_\text{inv}\)}} \\
        \cmidrule(lr){2-3}
        \cmidrule(lr){4-5}
        \midrule
        Txt
            & \(0.9_{\pm 0.1}\) & \textsuperscript{(5.0)}
            & \(28.4_{\pm 2.9}\) & \textsuperscript{(5.0)} \\
        Img
            & \(22.8_{\pm 4.2}\) & \textsuperscript{(4.0)}
            & \(56.3_{\pm 2.7}\) & \textsuperscript{(4.0)} \\
        Img + Txt
            & \sigA{\(29.8_{\pm 6.0}\)} & \sigA{\textsuperscript{(3.0)}}
            & \sigA{\(61.2_{\pm 4.2}\)} & \sigA{\textsuperscript{(3.0)}} \\
        Router
            & \sigA{\(55.6_{\pm 2.1}\)} & \sigA{\textsuperscript{(2.0)}}
            & \sigA{\(77.9_{\pm 1.7}\)} & \sigA{\textsuperscript{(1.0)}} \\
        Router\textsuperscript{\dag}
            & \sigA{\(59.0_{\pm 2.9}\)} & \sigA{\textsuperscript{(1.0)}}
            & \sigA{\(75.9_{\pm 2.6}\)} & \sigA{\textsuperscript{(2.0)}} \\
        \bottomrule
    \end{tabular}
\end{table}

%% file: tables/table6_mean_replacement.tex
\begin{table}[t]
\centering
\caption{\textbf{Mean Replacement.} Relative change in scores across five seeds when each sample's text embedding is replaced by the mean. Negative values indicate degradation. Bolded indicate \(p \leq 0.05\) on paired one-sided Wilcoxon signed-rank test.}
\label{tab:mean_replacement}
\begin{tabular}{llcc}
\toprule
dataset & annotation & \(\Delta F_1\) & \(\Delta\dinv\) \\
\midrule
BN-Sub \small(23 nodes)                & multi-label  & \(+1.5\%\) & \(-0.1\%\) \\
BN-Bio \small(293 nodes)               & multi-label  & \(+0.6\%\) & \(\mathbf{-21.7\%}\) \\
iNaturalist \small(10{,}000 endpoints) & single-label & \(\mathbf{-18.7\%}\) & \(\mathbf{-10.2\%}\) \\
\bottomrule
\end{tabular}
\end{table}

%% file: tables/table2_task_diagnostics.tex
\begin{table}[!htbp]
\centering
\caption{\textbf{VLM Inference Diagnostics.} Throughput, GPU memory allocation, and KV cache overhead for the variable proposal and captioning stages. Throughput is reported as mean ± std across batches. Dataset sizes differ between tasks due to filtering (variable proposal operates on the training partition and captioning operates on the full dataset).}
\label{tab:task_diagnostics}
\resizebox{\textwidth}{!}{%
\begin{tabular}{llccccc}
\toprule
    Task 
    & Dataset
    & \thead{Dataset Size \\ {[samples]}} 
    & \thead{Throughput \\ {[samples/s]}} 
    & \thead{Peak Alloc [GB] \\ (mean/max)} 
    & \thead{Peak Reserved [GB] \\ (mean/max)} 
    & \thead{KV Overhead \\ {[GB]}} \\
\midrule
    \multirow{3}{*}{Var. Proposal}
    & BN-Sub
        & 102,118 
        & 5.73 \(\pm\) 0.22 
        & 19.09 / 19.23 
        & 21.61 / 22.75 
        & 2.75 \\
    & BN-Bio 
        & 105,133 
        & 5.61 \(\pm\) 0.40 
        & 19.02 / 19.21 
        & 20.64 / 21.15 
        & 2.68 \\
    & FN 
        & 8,980 
        & 1.65 \(\pm\) 0.10 
        & 21.57 / 21.94 
        & 22.78 / 22.88 
        & 5.23 \\
\midrule
    \multirow{3}{*}{Captioning}
    & BN-Sub
        & 132,639 
        & 0.38 \(\pm\) 0.04 
        & 17.06 / 17.13 
        & 21.15 / 21.15 
        & 0.72 \\
    & BN-Bio
        & 142,637 
        & 0.29 \(\pm\) 0.01 
        & 17.04 / 17.12 
        & 21.15 / 21.15 
        & 0.70 \\
    & FN
        & 9,305 
        & 2.09 \(\pm\) 0.17 
        & 18.90 / 19.17 
        & 22.80 / 22.80 
        & 2.56 \\
\bottomrule
\end{tabular}%
}
\end{table}

%% file: tables/table3_training_diagnostics.tex
\begin{table}[!htbp]
\centering
\caption{\textbf{Training Model Diagnostics.} Trainable parameters, throughput (mean ± std across batches), and GPU memory usage for each model variant.}
\label{tab:training_diagnostics}
\resizebox{\textwidth}{!}{%
\begin{tabular}{llccccccc}
\toprule
    Dataset 
    & Model 
    & \thead{Num. \\ Classes} 
    & \thead{Batch \\ Size} 
    & \thead{Trainable \\ Params.} 
    & \thead{Throughput \\ {[samples/s]}} 
    & \thead{Model Mem \\ {[GB]}} 
    & \thead{Peak Alloc. [GB] \\ (mean/max)} 
    & \thead{Peak Reserved [GB] \\ (mean/max)} \\
\midrule
\multirow{5}{*}
    {BN-Sub} 
        & Txt 
            & 23 
            & 64 
            & 19M 
            & 985.96 \(\pm\) 50.77 
            & 0.09 
            & 2.57 / 2.62 
            & 3.81 / 3.88 \\
        & Img 
            & 23 
            & 64 
            & 19M 
            & 1,575.74 \(\pm\) 64.14 
            & 0.07 
            & 1.64 / 1.64 
            & 1.74 / 1.74 \\
        & Img + Txt 
            & 23 
            & 64 
            & 19M 
            & 1,281.58 \(\pm\) 59.64 
            & 0.09 
            & 1.86 / 1.87 
            & 2.11 / 2.11 \\
        & Router 
            & 23 
            & 64 
            & 10M 
            & 1,046.14 \(\pm\) 65.88 
            & 0.26 
            & 1.28 / 1.30 
            & 2.96 / 3.01 \\
        & Router\textsuperscript{\dag}
            & 23 
            & 64 
            & 66M 
            & 443.79 \(\pm\) 24.70 
            & 0.26 
            & 5.44 / 5.49 
            & 7.09 / 7.16 \\
\midrule
    \multirow{5}{*}{BN-Bio} 
        & Txt 
            & 293 
            & 64 
            & 19M 
            & 980.61 \(\pm\) 121.98 
            & 0.09 
            & 2.52 / 2.58 
            & 3.71 / 3.78 \\
        & Img 
            & 293 
            & 64 
            & 19M 
            & 1,596.01 \(\pm\) 134.24 
            & 0.09 
            & 1.65 / 1.65 
            & 1.76 / 1.76 \\
        & Img + Txt 
            & 293 
            & 64 
            & 19M 
            & 1,306.87 \(\pm\) 50.50 
            & 0.09 
            & 1.86 / 1.87 
            & 2.18 / 2.20 \\
        & Router 
            & 293 
            & 64 
            & 11M 
            & 1,053.14 \(\pm\) 44.21 
            & 0.28 
            & 1.28 / 1.30 
            & 2.91 / 2.96 \\
        & Router\textsuperscript{\dag} 
            & 293 
            & 64 
            & 69M 
            & 446.30 \(\pm\) 9.74 
            & 0.28 
            & 5.42 / 5.48 
            & 7.06 / 7.15 \\
\midrule
    \multirow{5}{*}{FN} 
        & Txt 
            & 192 
            & 64 
            & 19M 
            & 996.81 \(\pm\) 82.17 
            & 0.09 
            & 2.57 / 2.97 
            & 4.53 / 4.66 \\
        & Img 
            & 192 
            & 64 
            & 19M 
            & 1,634.52 \(\pm\) 51.52 
            & 0.07 
            & 1.65 / 1.65 
            & 1.74 / 1.74 \\
        & Img + Txt
            & 192 
            & 64 
            & 19M 
            & 1,310.38 \(\pm\) 39.97 
            & 0.09 
            & 1.87 / 1.94 
            & 2.15 / 2.16 \\
        & Router 
            & 192 
            & 64 
            & 11M 
            & 1,002.58 \(\pm\) 102.59 
            & 0.27
            & 1.29 / 1.43 
            & 3.56 / 3.65 \\
        & Router\textsuperscript{\dag} 
            & 192 
            & 64 
            & 68M 
            & 450.18 \(\pm\) 17.97 
            & 0.27 
            & 5.45 / 5.86 
            & 7.22 / 7.35 \\
\bottomrule
\end{tabular}%
}
\end{table}

%% file: tables/table4_hyperparameters.tex
\begin{table}[!htbp]
\centering
\caption{\textbf{Training Hyperparameters.} Learning rates and training epochs across models.}
\label{tab:training_hyperparams}
\begin{tabular}{llcc}
\toprule
    Dataset 
    & Model 
    & \thead{Learning \\ Rate} 
    & \thead{Num. \\ Epochs} \\
\midrule
\multirow{5}{*}{BN\textsuperscript{*}} 
    & Txt                          & 3e-5 & 35 \\
    & Img                          & 1e-4 & 35 \\
    & Img + Txt                    & 1e-4 & 35 \\
    & Router                       & 1e-4 & 35 \\
    & Router\textsuperscript{\dag} & 1e-4 & 35 \\
\midrule
\multirow{5}{*}{FN} 
    & Txt                          & 3e-5 & 35 \\
    & Img                          & 1e-4 & 15 \\
    & Img + Txt                    & 1e-4 & 15 \\
    & Router                       & 1e-4 &  8 \\
    & Router\textsuperscript{\dag} & 1e-4 &  8 \\
\bottomrule
\multicolumn{4}{l}{\footnotesize \textsuperscript{*}Applies to both BN-Sub and BN-Bio.} \\
\end{tabular}
\end{table}

%% file: tables/table5_embeddings_analysis.tex
\begin{table}[t]
\centering
\caption{\textbf{Complementarity of Frozen Embeddings.} Independence ratio \(P(\text{text correct}\mid\text{image wrong})\,/\,P(\text{text correct})\) by hierarchy level, from \(k\)-NN classifiers on the frozen image and text embeddings. iNaturalist level~1 is excluded (image accuracy \(99.6\%\)).}

\label{tab:emb_probe}
\begin{tabular}{lccccc}
\toprule
 & L1 & L2 & L3 & L4 & L5 \\
\midrule
BN-Sub      & 0.43 & 0.39 & 0.50 & --- & --- \\
BN-Bio      & 0.68 & 0.27 & 0.33 & 0.29 & --- \\
iNaturalist & \emph{excl.} & 0.61 & 0.63 & 0.64 & 0.58 \\
\bottomrule
\end{tabular}
\end{table}

%% file: tables/table11_caption_embeddings_compare.tex
\begin{table}[ht]
\centering
\caption{\textbf{Caption Sources.} Independence ratio (as in \autoref{tab:emb_probe}) on BN-Bio (OOD test, 20,000 samples) for five caption sources, each embedded with Qwen3-Embedding-8B. Lower values indicate greater redundancy with the image embedding.}
\label{tab:caption_probe}
\begin{tabular}{lcccc}
\toprule
Captions & L1 & L2 & L3 & L4 \\
\midrule
Elicited (ours)        & 0.68 & 0.27 & 0.33 & 0.29 \\
Keyword prose (ours)   & 0.64 & 0.27 & 0.33 & 0.32 \\
Description            & 0.48 & 0.27 & 0.34 & 0.33 \\
Visual                 & 0.41 & 0.27 & 0.33 & 0.28 \\
Plain                  & 0.36 & 0.26 & 0.32 & 0.27 \\
\bottomrule
\end{tabular}
\end{table}

%% file: tables/table8_zero_shot_vlm.tex
\begin{table}[t]
\centering
\caption{\textbf{Zero-Shot VLM Classification on BenthicNet.} Scheme conformance and prediction cardinality for free-form prompting and for prompting conditioned on the classification scheme.}
\label{tab:zero_shot_vlm}
\begin{tabular}{lcc}
\toprule
 & BN-Sub & BN-Bio \\
\midrule
images yielding no scheme-conformant term (free-form) & \(92.0\%\) & \(77.6\%\) \\
conformance when conditioned on the scheme            & \(99.9\%\) & \(98.0\%\) \\
predicted / annotated labels per image (conditioned) & \(2.08\) & \(6.98\) \\
\bottomrule
\end{tabular}
\end{table}

%% file: tables/table10_noise_ablation.tex
\begin{table}[t]
\centering
\caption{\textbf{Scale-Matched Noise Ablation.} Median scores across seeds for the Router with clean text embeddings and retrained with scale-matched Gaussian noise in their place; \(p\) is a two-sided Wilcoxon signed-rank test over paired seeds. Since iNat has only five seeds, the lowest attainable \(p\)-value is 0.063.}
\label{tab:noise_ablation}
\begin{tabular}{lrcccccc}
\toprule
 & & \multicolumn{3}{c}{Macro \(F_1\)} & \multicolumn{3}{c}{Macro \(\dinv\)} \\
\cmidrule(lr){3-5}\cmidrule(lr){6-8}
dataset & seeds & clean & noise & \(p\) & clean & noise & \(p\) \\
\midrule
BN-Sub      & 10 & 0.678 & 0.241 & 0.002 & 0.654 & 0.006 & 0.002 \\
BN-Bio      & 10 & 0.129 & 0.047 & 0.002 & 0.037 & 0.001 & 0.002 \\
FathomNet   & 30 & 0.410 & 0.414 & 0.670 & 0.323 & 0.316 & 0.008 \\
iNaturalist &  5 & 0.549 & 0.190 & 0.063 & 0.770 & 0.388 & 0.063 \\
\bottomrule
\end{tabular}
\end{table}

%% file: tables/table7_dclip.tex
\begin{table}[t]
\centering
\caption{\textbf{Descriptor Matching on BN-Bio.} Macro \(F_1\) and macro \(\dinv\) (\%) of a DCLIP-style baseline \citep{dclip}, with and without the hierarchical constraint \(\fCM\), alongside the trained models. Descriptors are embedded with the CLIP ViT-L-14-quickgelu text encoder. DCLIP rows are single deterministic runs; trained rows are the ten-seed OOD values of \autoref{tab:results}.}
\label{tab:dclip}
\begin{tabular}{lcc}
\toprule
 & Macro \(F_1\) & Macro \(\dinv\) \\
\midrule
DCLIP, no constraint       & 5.8  & 0.12 \\
DCLIP \(+\ \fCM\)            & 5.8  & 0.17 \\
\midrule
Text-only (trained)        & 7.4  & 2.5 \\
Image-only (trained)       & 11.4 & 4.4 \\
Router (trained)           & 13.2 & 5.9 \\
Router\(^\dagger\) (trained) & 14.3 & 6.4 \\
\bottomrule
\end{tabular}
\end{table}

%% file: figures/fig4_text_tow/fig4_text_tow.tex
\begin{figure}[!htbp]
    \centering
    \begin{subfigure}[b]{0.49\textwidth}
        \centering
        \includegraphics[width=\textwidth]{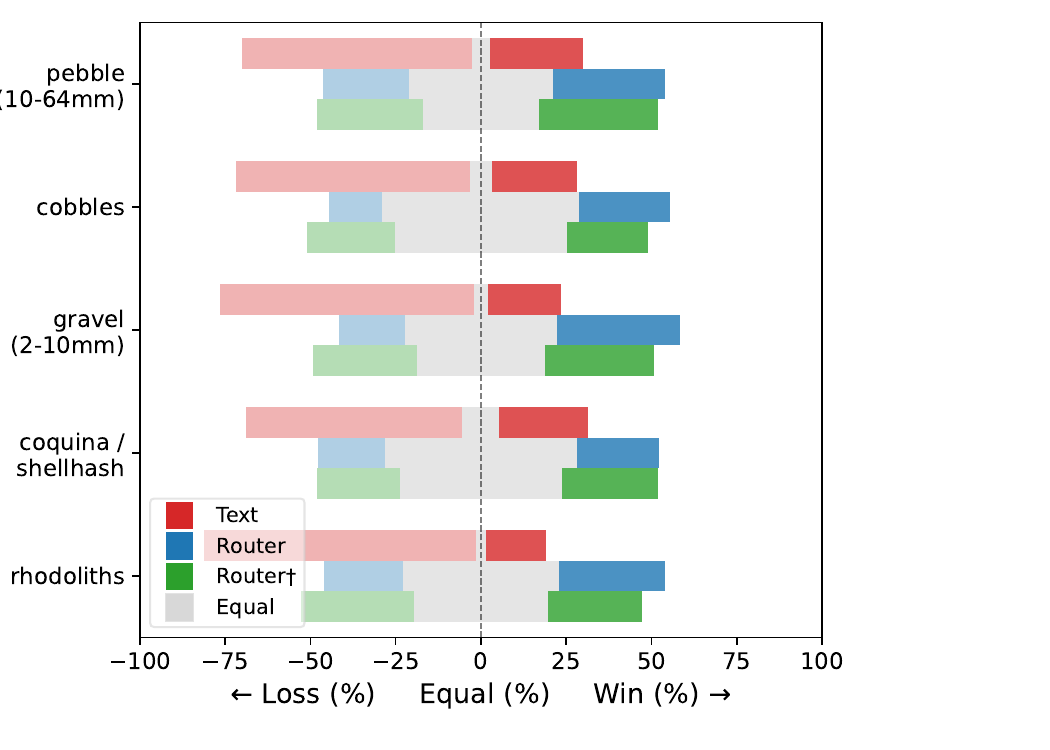}
        \caption{BN-Sub (OOD)}
        \label{fig:bn_sub_text_tow}
    \end{subfigure}
    \begin{subfigure}[b]{0.49\textwidth}
        \centering
        \includegraphics[width=\textwidth]{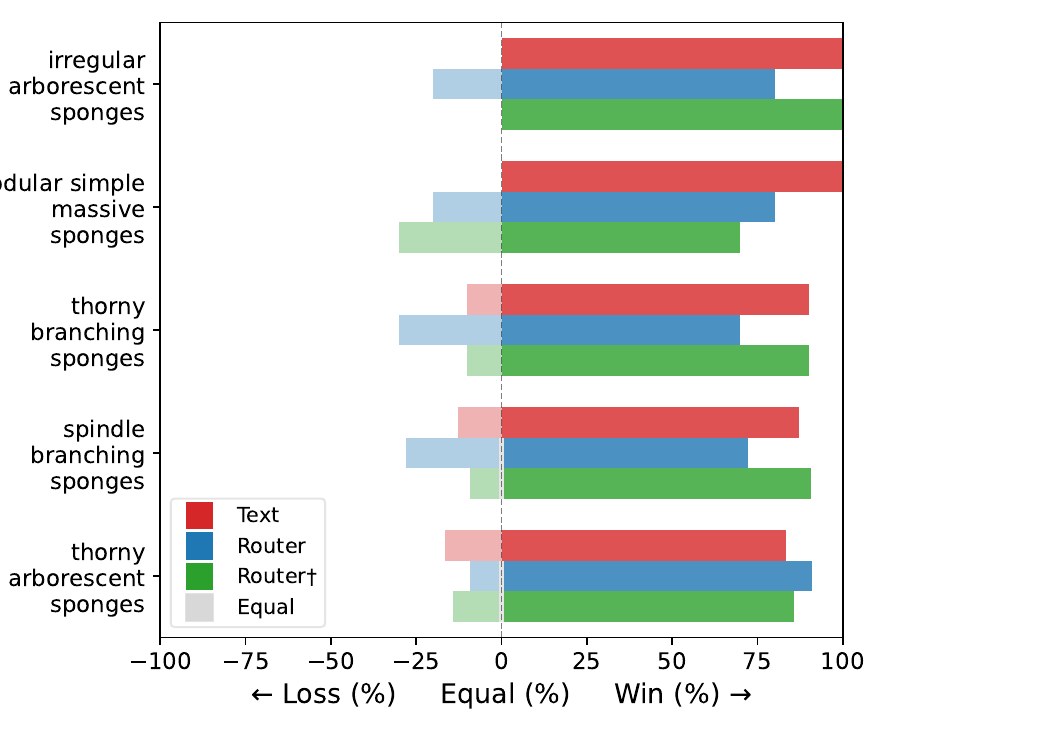}
        \caption{BN-Bio (OOD)}
        \label{fig:bn_bio_text_tow}
    \end{subfigure}
    \caption{\textbf{Modality Information Exclusivity.} Displayed are the top-5 nodes in which text holds the greatest advantages over vision in the BenthicNet datasets in terms of \(d_\text{inv}\). The bars depict the percentage of samples for each node in which the models perform less than, equal to, or greater than vision.}
    \label{fig:text_tow}
\end{figure}

%% file: figures/fig4-5_property_correlations/fig4-5_property_correlations.tex
\begin{figure}[!htbp]
    \centering
    \begin{subfigure}[b]{0.75\textwidth}
        \centering
        \includegraphics[width=\textwidth]{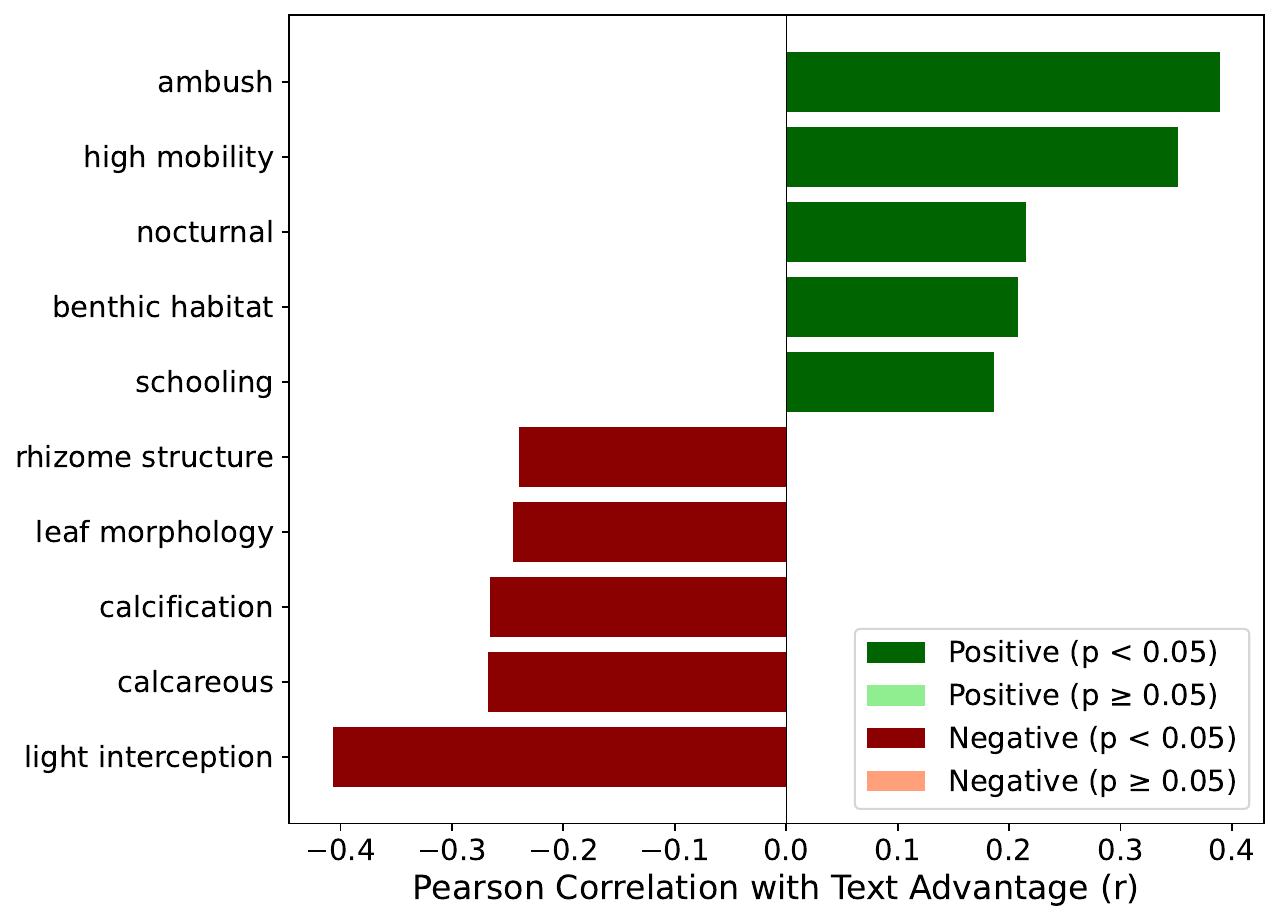}
        \phantomcaption
        \label{fig:property_correlation_bars}
    \end{subfigure}
    \begin{subfigure}[b]{0.98\textwidth}
        \centering
        \includegraphics[width=\textwidth]{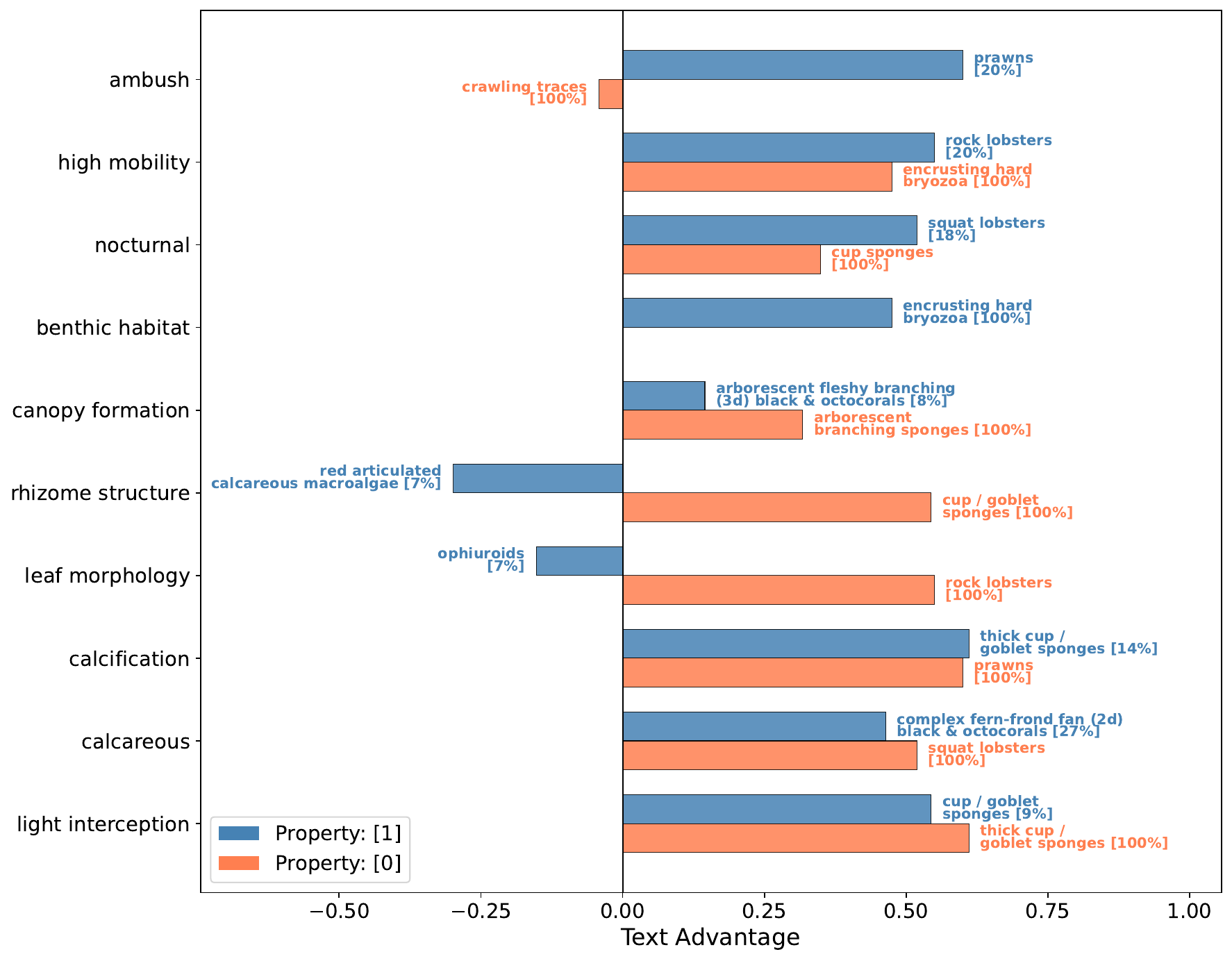}
        \phantomcaption
        \label{fig:property_class_pairs}
    \end{subfigure}
    \caption{\textbf{Variable Correlations with Text Modality Advantage.}
    (a)~Pearson correlation between each variable's positive annotation in captions and text advantage across all nodes. Dark shading indicates statistical significance (\(p < 0.05\)).
    (b)~For each top-correlated variable (sorted by \(r\)), we select a representative class where the variable benefits text when it is positive [1] (blue) and one where it is negative [0]. Sample coverage is noted in brackets. A soft diversity penalty discourages repeated selection of the same class across properties. Since the majority of samples have a positive annotation for ``benthic habitat'', it is the only variable without a negative bar.}
    \label{fig:property_correlations}
\end{figure}

%% file: tables/table9_gate_weights.tex
\begin{table}[t]
\centering
\caption{\textbf{Gate Weights under Text Ablation on BN-Bio.} Text gate weight \(w_{\mathrm{txt}}\) with clean captions and with every caption replaced by the mean embedding, the relative change, and the number of seeds in which the value decreases.}
\label{tab:gate}
\begin{tabular}{lcccc}
\toprule
 & clean & constant text & \(\Delta\) & seeds lower \\
\midrule
mean \(w_{\mathrm{txt}}\)               & 0.5658 & 0.5455 & \(-3.6\%\)  & 5/5 \\
per-sample std.\ of \(w_{\mathrm{txt}}\) & 0.1315 & 0.1306 & \(-0.68\%\) & --- \\
\bottomrule
\end{tabular}
\end{table}

%% file: algorithms/algorithm2a_random_hierarchy_generation.tex
\begin{algorithm}[htbp]
\caption{Random Hierarchy Generation}\label{alg:hierarchy_gen}
\begin{algorithmic}[1]
\Require Number of nodes \(|\mathcal{V}|\)
\Ensure Descendant matrix \(\mathbf{D} \in \{0,1\}^{|\mathcal{V}| \times |\mathcal{V}|}\)

\State Select number of root nodes \(n_r \in [1, \max(1, \lfloor |\mathcal{V}| / 100 \rfloor)]\) uniformly at random
\State Initialise a forest with root nodes \(v = 0, \ldots, n_r - 1\)

\For{each remaining node \(v = n_r, \ldots, |\mathcal{V}| - 1\)}
    \State Assign \(v\) as a child of a node chosen uniformly at random from \(\{0, \ldots, v - 1\}\)
\EndFor

\State \(\mathbf{D} \gets \mathbf{0}^{|\mathcal{V}| \times |\mathcal{V}|}\)

\For{each node \(v \in \mathcal{V}\)}
    \State Set \(D_{v,u} \gets 1\) for all descendants \(u\) of \(v\) (including \(v\) itself)
\EndFor

\State \Return \(\mathbf{D}\)
\end{algorithmic}
\end{algorithm}

%% file: algorithms/algorithm2b_random_hierarchical_data_generation.tex
\begin{algorithm}[t]
\caption{Random Hierarchical Data Generation}\label{alg:data_gen}
\begin{algorithmic}[1]
\Require Number of nodes \(|\mathcal{V}|\), number of samples \(N\), descendant matrix \(\mathbf{D}\), sparsity \(s\)
\Ensure Data matrix \(\mathbf{Y} \in \{0,1\}^{N \times |\mathcal{V}|}\)

\State Define neighbourhood \(\delta = \max(1, \lfloor s / 2 \rfloor)\)
\State \(\mathbf{Y} \gets \mathbf{0}^{N \times |\mathcal{V}|}\)

\For{each sample \(i = 1, \ldots, N\)}
    \State Sample \(k\) uniformly from \([\max(1, s - \delta),\; \min(|\mathcal{V}|, s + \delta)]\)
    \State Select \(k\) nodes uniformly at random from \(\mathcal{V}\)
    \State Activate selected nodes: \(\mathbf{r} \in \{0,1\}^{|\mathcal{V}|}\)
    \State Propagate through hierarchy: \(\mathbf{Y}_i \gets \text{clip}(\mathbf{D} \cdot \mathbf{r},\; 0,\; 1)\) \Comment{Enforce ancestor activation}
\EndFor

\State \Return \(\mathbf{Y}\)
\end{algorithmic}
\end{algorithm}

%% file: algorithms/algorithm2c_realistic_hml_noise_generation.tex
\begin{algorithm}[htbp]
\caption{Hierarchical Noise Generation}\label{alg:noise_gen}
\begin{algorithmic}[1]
\Require Ground-truth matrix \(\mathbf{Y} \in \{0,1\}^{N \times |\mathcal{V}|}\), descendant matrix \(\mathbf{D}\), noise level \(\eta \in [0,1]\)
\Ensure Noisy predictions \(\hat{\mathbf{Y}}\), prediction scores \(\hat{\mathbf{S}}\)

\State Compute depth \(\ell_v \gets \sum_{u} D_{u,v} - 1\) and normalised depth \(d_v \gets \ell_v / \max_{u} \ell_u\) for each node \(v \in \mathcal{V}\)
\State Define FN rate per node: \(r_v^{\text{FN}} = 0.5\,\eta\, d_v^{1.5}\) \Comment{Increases with depth}
\State Define FP rate per node: \(r_v^{\text{FP}} = 10^{-4}\,\eta\,\bigl(1 + 0.5\,(1 - d_v)\bigr)\) \Comment{Higher near root}

\State \(\hat{\mathbf{Y}} \gets \mathbf{Y}\) \Comment{Initialise predictions from ground truth}

\For{each node \(v \in \mathcal{V}\)}
    \For{each sample \(i\) where \(Y_{iv} = 1\)}
        \State Flip \(\hat{Y}_{iv} \gets 0\) with probability \(r_v^{\text{FN}}\) \Comment{Inject false negatives}
    \EndFor
    \For{each sample \(i\) where \(Y_{iv} = 0\)}
        \State Flip \(\hat{Y}_{iv} \gets 1\) with probability \(r_v^{\text{FP}}\) \Comment{Inject false positives}
    \EndFor
\EndFor

\For{each sample \(i = 1, \ldots, N\)} \Comment{Enforce hierarchical consistency}
    \State \(\hat{\mathbf{Y}}_i \gets \text{clip}(\mathbf{D} \cdot \hat{\mathbf{Y}}_i,\; 0,\; 1)\)
\EndFor

\For{each sample \(i\) and node \(v \in \mathcal{V}\)} \Comment{Scores for AP}
    \If{\(\hat{Y}_{iv} = 1\)}
        \State \(\hat{S}_{iv} \sim \mathcal{U}(0.55, 0.95)\)
    \Else
        \State \(\hat{S}_{iv} \sim \mathcal{U}(0.05, 0.45)\)
    \EndIf
\EndFor

\State \Return \(\hat{\mathbf{Y}}, \hat{\mathbf{S}}\)
\end{algorithmic}
\end{algorithm}

%% file: figures/fig5_metrics_comparison/fig5a_noise_effect.tex
\begin{figure}[t]
    \centering
    \begin{subfigure}[b]{0.48\textwidth}
        \centering
        \includegraphics[width=\textwidth]{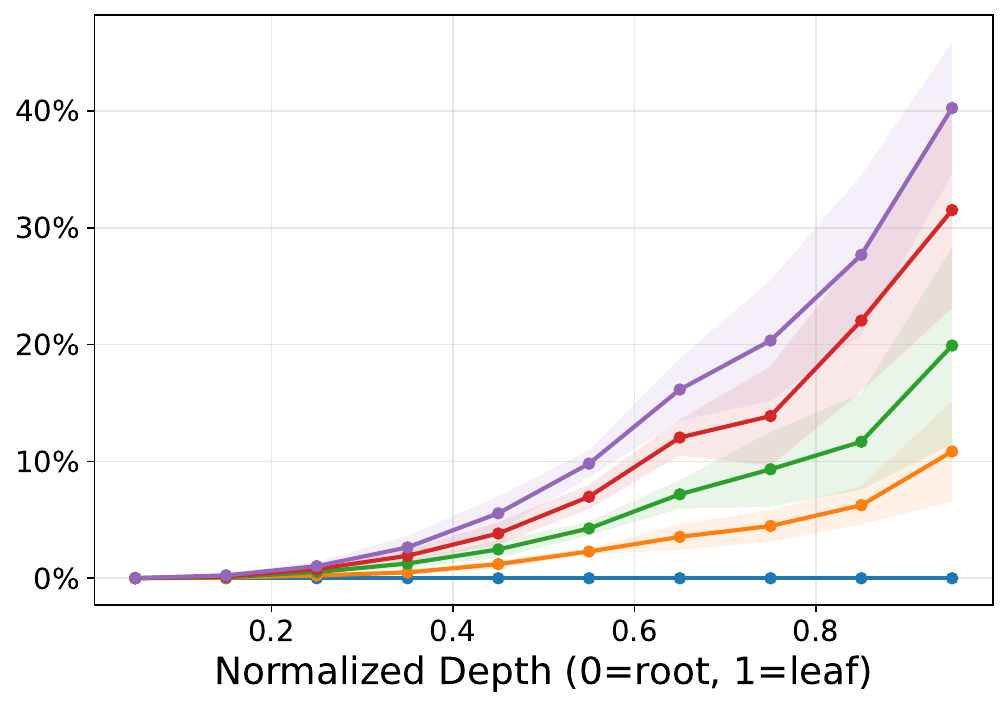}
        \caption{False Negative (FN) Rate}
        \label{fig:noise_experiment_fn_rate}
    \end{subfigure}
    \hfill
    \begin{subfigure}[b]{0.48\textwidth}
        \centering
        \includegraphics[width=\textwidth]{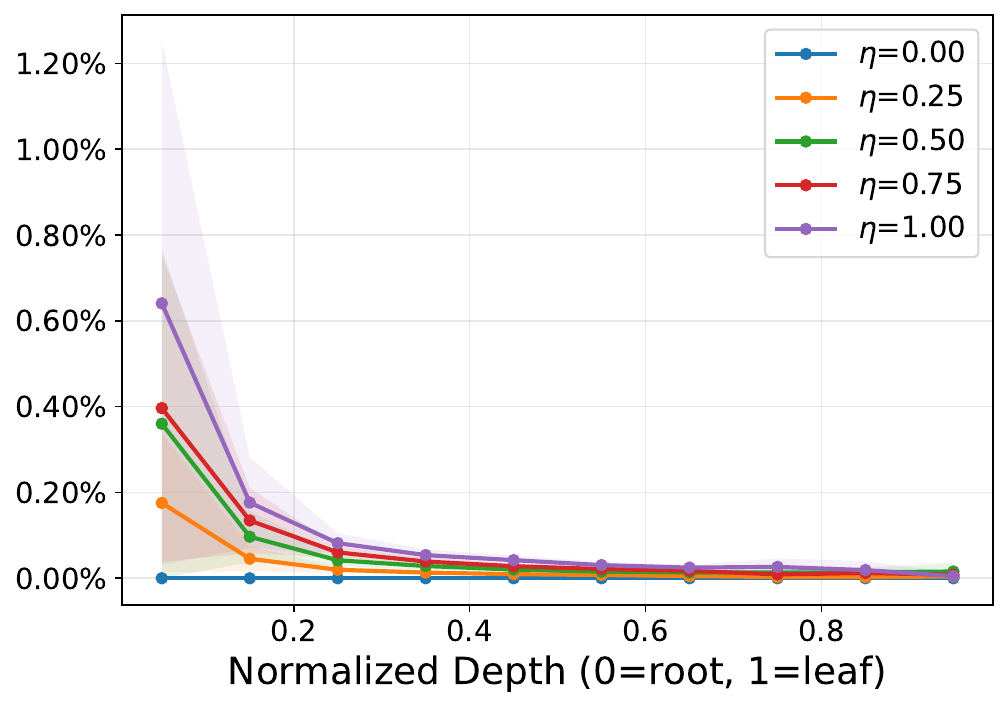}
        \caption{False Positive (FP) Rate}
        \label{fig:noise_experiment_fp_rate}
    \end{subfigure}
    \caption{\textbf{Random Noise Profile.} The behaviour of the random noise added is such that it slightly increases the FP rate at higher-level nodes, while substantially increasing the FN rate at deeper nodes, simulating a more realistic type of HML model degradation.}
    \label{fig:noise_experiment_profile}
\end{figure}

%% file: figures/fig5_metrics_comparison/fig5b_metrics_results.tex
\begin{figure}[t]
    \centering
    \begin{subfigure}[b]{0.32\textwidth}
        \centering
        \includegraphics[width=\textwidth]{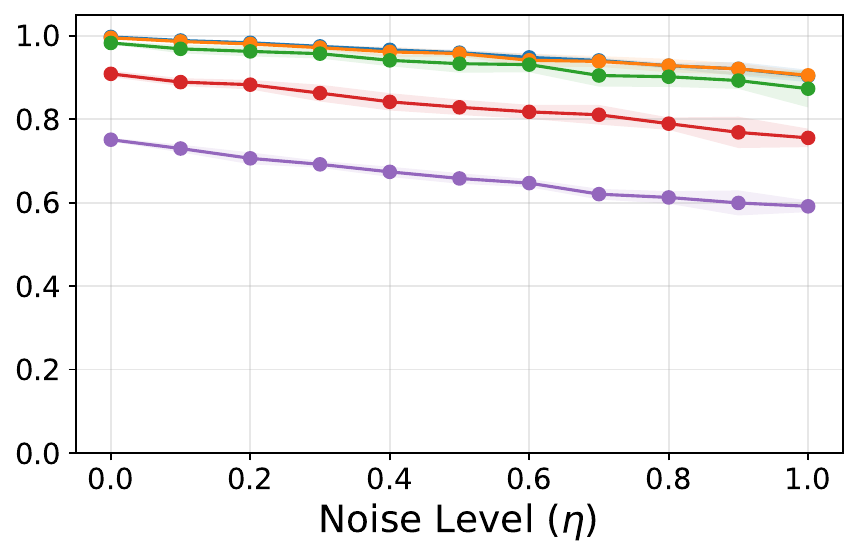}
        \caption{\(F_\text{1}\)}
        \label{fig:noise_experiment_macro_f1}
    \end{subfigure}
    \hfill
    \begin{subfigure}[b]{0.32\textwidth}
        \centering
        \includegraphics[width=\textwidth]{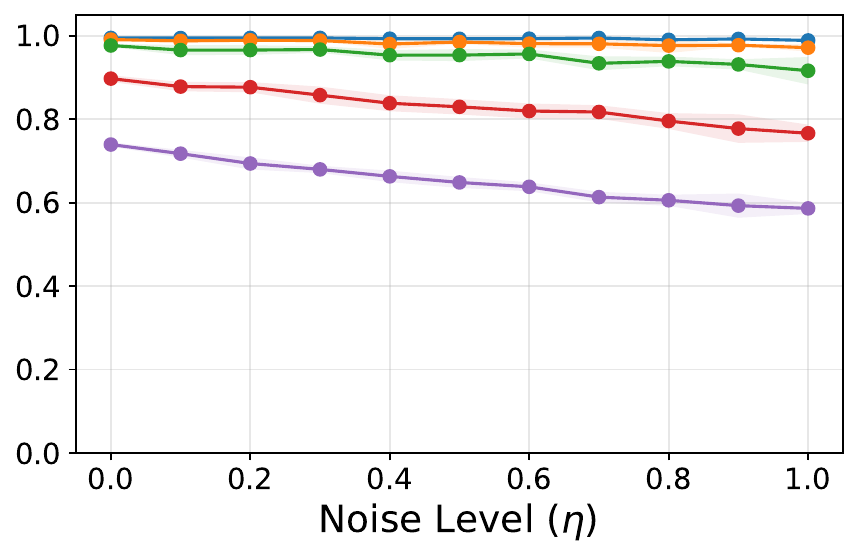}
        \caption{Precision}
        \label{fig:noise_experiment_precision}
    \end{subfigure}
    \hfill
    \begin{subfigure}[b]{0.32\textwidth}
        \centering
        \includegraphics[width=\textwidth]{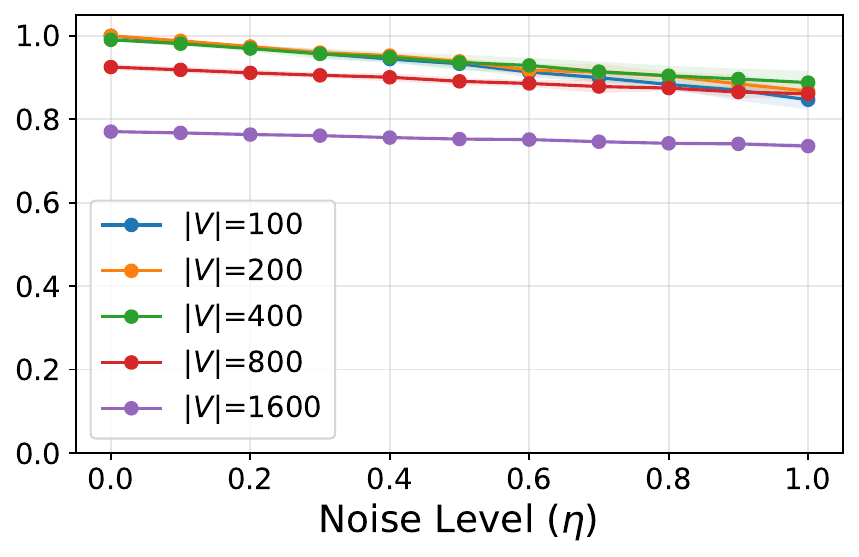}
        \caption{Recall}
        \label{fig:noise_experiment_recall}
    \end{subfigure}

    \vspace{1em}

    \begin{subfigure}[b]{0.32\textwidth}
        \centering
        \includegraphics[width=\textwidth]{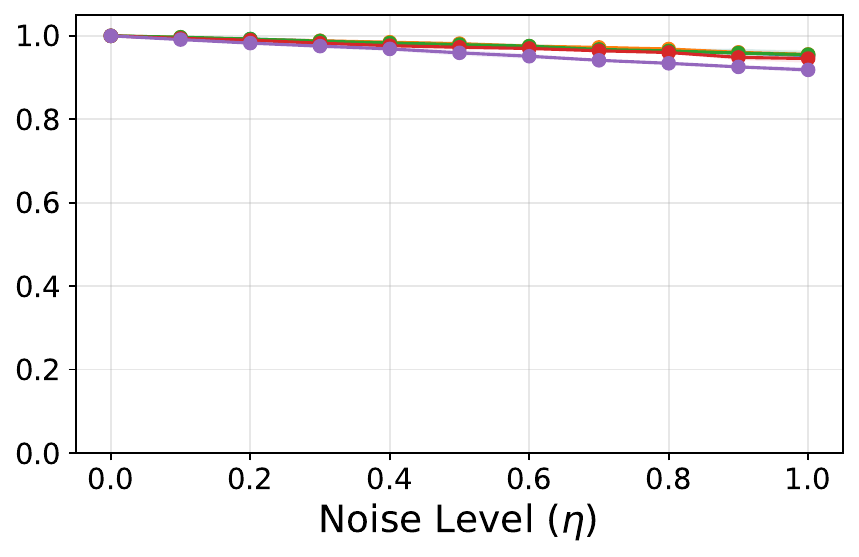}
        \caption{AP}
        \label{fig:noise_experiment_ap}
    \end{subfigure}
    \begin{subfigure}[b]{0.32\textwidth}
        \centering
        \includegraphics[width=\textwidth]{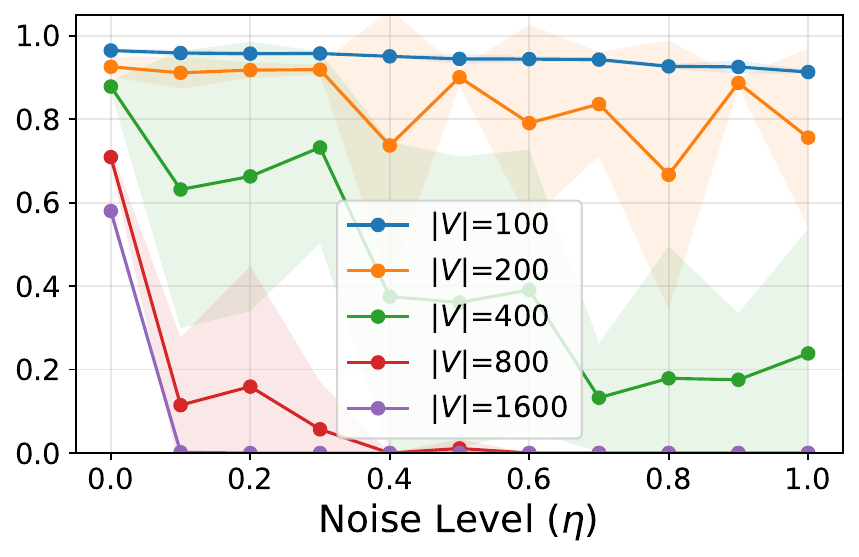}
        \caption{\(d_{\text{inv}}\)}
        \label{fig:noise_experiment_inv_hml}
    \end{subfigure}
    \hfill

    \caption{\textbf{HML Metric Behaviours.} The top row displays the behaviour of node-wise \(F_\text{1}\), and its components in the simulated noisy model experiment across hierarchy sizes \(|\mathcal{ V}|\) and added \(\eta\). The bottom row compares AP to \(d_\text{inv}\).}
    \label{fig:noise_experiment_metrics_comparison}
\end{figure}

%% file: figures/fig6_hf1_d_inv_hierarchy/fig6_hf1_d_inv_hierarchy.tex
\begin{figure}[t]
    \centering
    \includegraphics[width=\linewidth]{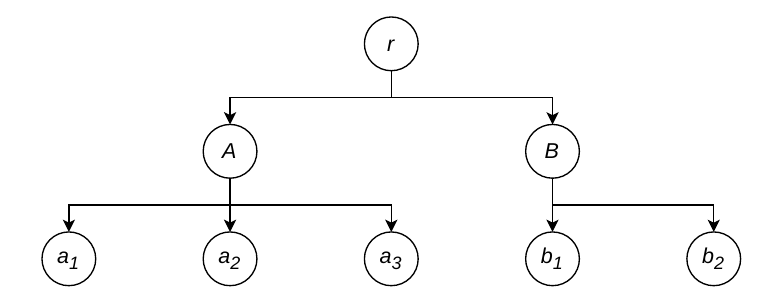}
    \caption{\textbf{Sample Hierarchy.} A toy hierarchy of eight nodes, intended as the subject for comparisons between \(hF_\text{1}\) and \(d_\text{inv}\).}
    \label{fig:hf1_comparison}
\end{figure}

%% file: tables/table6-5_decomposition.tex
\begin{table}[t]
\centering
\caption{\textbf{Decomposition of the \(\dinv\) Change under Mean Replacement.} BN-Bio, five seeds (\autoref{tab:mean_replacement}), rescored with alternative cardinality factors on the same matched predictions; setting \(\gamma \equiv 1\) leaves only the matched proximity \(s_{\max}\). Per-node: nodes that degrade out of nodes evaluated, with \(p\)-value. Without the text signal, mean predicted endpoints rise from \(10.16\) to \(10.29\) (ground truth: \(2.34\)).}
\label{tab:decomposition}
\begin{tabular}{lcccl}
\toprule
score & clean & ablated & \(\Delta\) & per-node \\
\midrule
\(\dinv\), \(k_P=0.5\)        & 0.0391 & 0.0306 & \(\mathbf{-21.7\%}\) & \(214/248\), \(1.6\times10^{-32}\) \\
min/max ratio             & 0.2553 & 0.2452 & \(-4.0\%\) & \(193/248\), \(1.1\times10^{-21}\) \\
\(F_1\)-style count penalty & 0.3695 & 0.3594 & \(-2.7\%\) & \(183/248\), \(2.8\times10^{-16}\) \\
\(hF_1\)                    & 0.3857 & 0.3811 & \(-1.2\%\) & \(158/246\), \(4.7\times10^{-6}\) \\
\midrule
proximity, \(\gamma=1\)     & 0.8371 & 0.8408 & \(+0.43\%\) & \(60/246\), \(p=1.0\) \\
\bottomrule
\end{tabular}
\end{table}

%% file: theorems/thm1_proof_of_kp_star.tex
\begin{theorem}[Critical Value for the Trivial Root of \(g\)]
\label{thm:proof_of_kp_star}
Given \(n_{\text{ME}} > 1\) and \(k_{\text{P}} \in (0,1)\), then at  \(k_{\text{P}} = k^{*}_{\text{P}} := \frac{n_{\text{ME}}-1}{n_{\text{ME}}}\), the only root of:
\begin{equation}
\label{eq:func_form_of_f_of_b}
    g(b) = (k_{\text{P}}-1)e^{-bn_{\text{ME}}}+e^{-b}-k_{\text{P}},
\end{equation} is the trivial solution \(b=0\).
\end{theorem}

\begin{proof}
We begin by observing that \(b=0\) is always a root of \(g\),
since \(g(0) = (k_{\text{P}}-1) + 1 - k_{\text{P}} = 0\) for
every \(k_{\text{P}}\).

We then compute the first derivative of \(g(b)\) at \(b=0\):
\begin{align}
    \label{eq:first_derivative_g}
    &\diff{g}{b} = -n_{\text{ME}}(k_{\text{P}}-1)e^{-bn_{\text{ME}}}-e^{-b}
    \\
    &\diff*{g}{b}{b=0} = -n_{\text{ME}}(k_{\text{P}}-1)-1
\end{align}
Setting \(\diff*{g}{b}{b=0}=0\), we have:
\begin{equation}
    \label{eq:kp_at_local_max}
    k^{*}_{\text{P}} = \frac{n_{\text{ME}}-1}{n_{\text{ME}}},
\end{equation}
as the form \(k^{*}_{\text{P}}\) must take for \(b=0\) to be a
critical point of \(g\). Plugging this back into \(\diff{g}{b}\),
we can observe the behaviour of the function to the left and right
of \(b=0\):
\begin{align}
    \diff{g}{b} &= n_{\text{ME}}\frac{1}{n_{\text{ME}}}
    e^{-bn_{\text{ME}}}-e^{-b} \\
    &= e^{-bn_{\text{ME}}}-e^{-b}.
\end{align}
Since \(n_{\text{ME}} > 1\), for \(b > 0\) we have
\(bn_{\text{ME}} > b\) and hence
\(e^{-bn_{\text{ME}}} < e^{-b}\), giving
\(\diff{g}{b} < 0\). Symmetrically, for \(b < 0\) we have
\(bn_{\text{ME}} < b\) and hence
\(e^{-bn_{\text{ME}}} > e^{-b}\), giving
\(\diff{g}{b} > 0\). This means that for \(k^{*}_{\text{P}}\),
\(b=0\) is a maximum, and the function monotonically increases on
its left to meet this value and decreases monotonically on its
right.

For completeness, we can examine the bounds of the function.
It is observable from the form presented in
\autoref{eq:func_form_of_f_of_b} that:
\begin{equation}
    \lim_{b \to \infty} g(b) = 0 + 0 - k^{*}_{\text{P}}
    = -k^{*}_{\text{P}} < 0.
\end{equation}
For the other direction, we factor \(g\) as:
\begin{equation}
    g(b) = e^{-b}\big[(k_{\text{P}}-1)e^{-b(n_{\text{ME}}-1)}+1\big]
    -k_{\text{P}}.
\end{equation}
As \(b \to -\infty\), the prefactor \(e^{-b} \to +\infty\).
Inside the bracket, \(e^{-b(n_{\text{ME}}-1)} \to +\infty\)
(since \(n_{\text{ME}}-1 > 0\)), and its coefficient
\((k_{\text{P}}-1) < 0\) drives the bracket to \(-\infty\).
The product of \(e^{-b} \to +\infty\) with a bracket tending
to \(-\infty\) gives:
\begin{equation}
    \lim_{b \to -\infty} g(b) = -\infty.
\end{equation}
\noindent
Therefore, for \(k^{*}_{\text{P}}\), \(b=0\) is the global
maximum with \(g(0)=0\), so \(g(b) \leq 0\) for all
\(b \in \mathbb{R}\) with equality only at \(b=0\). Thus,
\(b=0\) is the only root of \(g(b)\).
\end{proof}

%% file: theorems/thm2_proof_of_lin_for_kp_star.tex
\begin{theorem}
\label{thm:proof_kp_star_lin}
At \(k_{\text{P}} = k^{*}_{\text{P}}\), the exponential decay function
\(\gamma(n_{\text{E}};k_{\text{P}},n_{\text{ME}}) = a e^{-b n_{\text{E}}} + c\)
reduces to a linear function of \(n_{\text{E}}\).
\end{theorem}

\begin{proof}
From \autoref{thm:proof_of_kp_star}, at
\(k_{\text{P}} = k^{*}_{\text{P}}\) the only solution is \(b=0\),
so it suffices to evaluate
\(\lim_{b \to 0} \gamma(n_{\text{E}};k_{\text{P}},n_{\text{ME}})\).

Returning to the exponential decay function and plugging in the
expressions for \(a\) and \(c\) derived earlier,
namely \(a = (1-e^{-bn_{\text{ME}}})^{-1}\) and \(c = 1-a\):
\begin{align}
    \gamma(n_{\text{E}};k_{\text{P}},n_{\text{ME}})
    &= \frac{e^{-b n_{\text{E}}}}{1-e^{-bn_{\text{ME}}}}
       + 1-\frac{1}{1-e^{-bn_{\text{ME}}}} \\
    &= 1-\frac{1-e^{-bn_{\text{E}}}}{1-e^{-bn_{\text{ME}}}}.
\end{align}
Now, we can perform a Taylor expansion of the exponential terms
around \(b=0\):
\begin{align}
    \gamma(n_{\text{E}};k_{\text{P}},n_{\text{ME}})
    &= 1-\frac{1-\big(1-bn_{\text{E}}
       +\mathcal{O}(b^2)\big)}{1-\big(1-bn_{\text{ME}}
       +\mathcal{O}(b^2)\big)} \\
    &= 1-\frac{bn_{\text{E}}
       +\mathcal{O}(b^2)}{bn_{\text{ME}}
       +\mathcal{O}(b^2)} \\
    &= 1-\frac{n_{\text{E}}
       +\mathcal{O}(b)}{n_{\text{ME}}+\mathcal{O}(b)}.
\end{align}
Therefore, we have:
\begin{equation}
    \lim_{b \to 0}
    \gamma(n_{\text{E}};k_{\text{P}},n_{\text{ME}})
    = 1-\frac{n_{\text{E}}}{n_{\text{ME}}},
\end{equation}
which is linear in \(n_{\text{E}}\). As a final note, plugging
in \(n_{\text{E}}=1\) into this linear form gives
\(\frac{n_{\text{ME}}-1}{n_{\text{ME}}}\), which is the form for
\(k^{*}_{\text{P}}\) derived in \autoref{eq:kp_at_local_max},
providing a consistency check.
\end{proof}

%% file: theorems/thm3_proof_of_b_exist.tex
\begin{theorem}[Non-Trivial Existence of \(b\)]
\label{thm:proof_nontrivial_b}
For a given \(n_{\text{ME}} > 1\) and
\(k_{\text{P}} \in (0,1) \setminus \{k^{*}_{\text{P}}\}\),
there exists for \(g(b)\) a unique non-trivial root. Moreover,
this root is positive when \(k_{\text{P}} < k^{*}_{\text{P}}\)
and negative when \(k_{\text{P}} > k^{*}_{\text{P}}\).
\end{theorem}

\begin{proof}
We return to our derivative for \(g(b)\) from
\autoref{eq:first_derivative_g}, which we restate and set equal
to zero:
\begin{equation}
    -n_{\text{ME}}(k_{\text{P}}-1)e^{-bn_{\text{ME}}}-e^{-b} = 0.
\end{equation}
In this form, we are now able to isolate for \(b\), attaining:
\begin{align}
    &-n_{\text{ME}}(k_{\text{P}}-1)e^{-bn_{\text{ME}}} = e^{-b} \\
    \label{eq:exponential_growth_for_bmax}
    &\implies n_{\text{ME}}(1-k_{\text{P}}) =
    e^{b(n_{\text{ME}}-1)} \\
    \label{eq:b_crit_identity}
    &\therefore b_{\text{crit}} =
    \frac{\ln\big(n_{\text{ME}}(1-k_{\text{P}})\big)}
    {n_{\text{ME}}-1}.
\end{align}
We note that the exponential term in the intermediate
\autoref{eq:exponential_growth_for_bmax} is strictly
monotonically increasing in the domain
\((-\infty, \infty)\); therefore, \(b_{\text{crit}}\)
must be the unique intersection with the constant on the
left-hand side, and hence the unique critical point of \(g\).
As a quick sanity check, we can observe that
\(b_{\text{crit}} \to 0\) as
\(k_{\text{P}} \to k^{*}_{\text{P}}\).

Now, plugging in \(b_{\text{crit}}\) in for the second
derivative of \(g(b)\), we get:
\begin{align}
    \diff*[2]{g}{b}{b=b_{\text{crit}}}
    &= n_{\text{ME}}^2(k_{\text{P}}-1)
       e^{-b_{\text{crit}}n_{\text{ME}}}
       +e^{-b_{\text{crit}}} \\
    &= n_{\text{ME}}^2(k_{\text{P}}-1)
       \big(n_{\text{ME}}(1-k_\text{P})\big)
       ^{-\frac{n_{\text{ME}}}{n_{\text{ME}}-1}}
       +\big(n_{\text{ME}}(1-k_\text{P})\big)
       ^{-\frac{1}{n_{\text{ME}}-1}} \\
    &= \big(n_{\text{ME}}(1-k_\text{P})\big)
       ^{-\frac{n_{\text{ME}}}{n_{\text{ME}}-1}}
       \big(n_{\text{ME}}(1-k_{\text{P}})
       +n_{\text{ME}}^2(k_\text{P}-1)\big) \\
    &= \big(n_{\text{ME}}(1-k_\text{P})\big)
       ^{-\frac{n_{\text{ME}}}{n_{\text{ME}}-1}}
       \big(n_{\text{ME}}(1-k_{\text{P}})
       -n_{\text{ME}}^2(1-k_\text{P})\big) \\
    &= \big(n_{\text{ME}}(1-k_\text{P})\big)
       ^{-\frac{n_{\text{ME}}}{n_{\text{ME}}-1}}
       \cdot(1-k_{\text{P}})\,n_\text{ME}(1-n_{\text{ME}}).
\end{align}
We see that all the terms above are positive except for
\(1-n_\text{ME}\), which confirms:
\begin{equation}
    \diff*[2]{g}{b}{b=b_{\text{crit}}} < 0,
\end{equation}
and \(b_{\text{crit}}\) as a maximum. Since
\(b_{\text{crit}}\) is also the only critical point and
\(g(b)\) is continuous, it is therefore the global maximum.

We can now plug \(b_{\text{crit}}\) back into the original
\(g(b)\):
\begin{align}
    g(b_{\text{crit}})
    &= (k_{\text{P}}-1)e^{-b_{\text{crit}}n_{\text{ME}}}
       +e^{-b_{\text{crit}}}-k_{\text{P}} \\
    &= (k_{\text{P}}-1)
       \big(n_{\text{ME}}(1-k_{\text{P}})\big)
       ^{-\frac{n_{\text{ME}}}{n_{\text{ME}}-1}}
       +\big(n_{\text{ME}}(1-k_{\text{P}})\big)
       ^{-\frac{1}{n_{\text{ME}}-1}}
       - k_{\text{P}} \\
    &= \big(n_{\text{ME}}(1-k_{\text{P}})\big)
       ^{-\frac{n_{\text{ME}}}{n_{\text{ME}}-1}}
       \big(n_{\text{ME}}(1-k_{\text{P}})
       -(1-k_{\text{P}})\big)-k_{\text{P}} \\
    \label{eq:intermediate_g_of_b_crit}
    &= \big(n_{\text{ME}}(1-k_{\text{P}})\big)
       ^{-\frac{n_{\text{ME}}}{n_{\text{ME}}-1}}
       (1-k_{\text{P}})(n_{\text{ME}}-1)-k_{\text{P}}.
\end{align}
When we insert \(k^{*}_{\text{P}}\) for \(k_{\text{P}}\),
keeping in mind that
\(1-k^{*}_{\text{P}}=\frac{1}{n_{\text{ME}}}\), it becomes
clear that the expression becomes zero.

We therefore proceed to look at the expression and
\(b_{\text{crit}}\) itself in the cases where
\(k_{\text{P}} < k^{*}_{\text{P}}\) and
\(k_{\text{P}} > k^{*}_{\text{P}}\). We rewrite the
expression in \autoref{eq:intermediate_g_of_b_crit} as:
\begin{align}
    g(b_{\text{crit}})
    &= n_{\text{ME}}^{-\frac{n_{\text{ME}}}{n_{\text{ME}}-1}}
       (1-k_{\text{P}})
       ^{1-\frac{n_{\text{ME}}}{n_{\text{ME}}-1}}
       (n_{\text{ME}}-1)-k_{\text{P}} \\
    &= n_{\text{ME}}^{-\frac{n_{\text{ME}}}{n_{\text{ME}}-1}}
       (1-k_{\text{P}})^{-\frac{1}{n_{\text{ME}}-1}}
       (n_{\text{ME}}-1)-k_{\text{P}}.
\end{align}
Now, to determine how \(g(b_{\text{crit}})\) behaves as
\(k_{\text{P}}\) changes, we take the derivative of
\(g(b_{\text{crit}})\) with respect to
\(k_{\text{P}}\)\footnote{Since \(b_{\text{crit}}\) has been
expressed in terms of \(k_{\text{P}}\) alone, this is a total
derivative.}:
\begin{align}
    \diff{g(b_{\text{crit}})}{k_{\text{P}}}
    &= -\frac{1}{n_{\text{ME}}-1}(n_{\text{ME}}-1)
       n_{\text{ME}}^{-\frac{n_{\text{ME}}}{n_{\text{ME}}-1}}
       (1-k_{\text{P}})
       ^{-\frac{n_{\text{ME}}}{n_{\text{ME}}-1}}(-1)-1 \\
    &= n_{\text{ME}}^{-\frac{n_{\text{ME}}}{n_{\text{ME}}-1}}
       (1-k_{\text{P}})
       ^{-\frac{n_{\text{ME}}}{n_{\text{ME}}-1}}-1 \\
    &= \big(n_{\text{ME}}(1-k_{\text{P}})\big)
       ^{-\frac{n_{\text{ME}}}{n_{\text{ME}}-1}}-1.
\end{align}
We can now attain the following statements:
\begin{align}
    \diff{g(b_{\text{crit}})}{k_{\text{P}}}=0
    &&\iff&& n_{\text{ME}}(1-k_{\text{P}})=1
    &&\iff&& &k_{\text{P}}=k^{*}_{\text{P}} \\
    \diff{g(b_{\text{crit}})}{k_{\text{P}}}<0
    &&\iff&& n_{\text{ME}}(1-k_{\text{P}})>1
    &&\iff&& &k_{\text{P}} < k^{*}_{\text{P}} \\
    \diff{g(b_{\text{crit}})}{k_{\text{P}}}>0
    &&\iff&& n_{\text{ME}}(1-k_{\text{P}})<1
    &&\iff&& &k_{\text{P}} > k^{*}_{\text{P}}.
\end{align}
It follows that \(g(b_{\text{crit}})\) attains a global
minimum at \(k^{*}_{\text{P}}\), where it equals zero: to
the left of \(k^{*}_{\text{P}}\), the function is descending
towards the minimum, while to the right, it is ascending away
from it. Therefore, \(g(b_{\text{crit}}) > 0\) for all
\(k_{\text{P}} \in (0,1) \setminus \{k^{*}_{\text{P}}\}\).

Next, returning to \autoref{eq:b_crit_identity}, we see that:
\begin{align}
    k_{\text{P}} < k^{*}_{\text{P}}
    &\implies 1-k_{\text{P}} > 1/n_{\text{ME}} \\
    &\implies n_{\text{ME}}(1-k_{\text{P}}) > 1 \\
    &\implies \ln\big(n_{\text{ME}}(1-k_{\text{P}})\big) > 0 \\
    &\therefore b_{\text{crit}} > 0.
\end{align}
If we apply the same logic to the case where
\(k_{\text{P}} > k^{*}_{\text{P}}\), we get the opposite
result, where \(b_{\text{crit}} < 0\).

To summarise, we now have the following relevant information:
\begin{enumerate}
    \item \(b_{\text{crit}}\) is the unique critical point
    of \(g(b)\) and a global maximum, since \(g(b)\) is
    continuous with no other critical points;
    \item \(b_{\text{crit}}\) is positive for
    \(k_{\text{P}}<k^{*}_{\text{P}}\), negative for
    \(k_{\text{P}}>k^{*}_{\text{P}}\), and zero for
    \(k_{\text{P}}=k^{*}_{\text{P}}\);
    \item \(g(b_{\text{crit}}) > 0\) for all
    \(k_{\text{P}} \in (0,1) \setminus
    \{k^{*}_{\text{P}}\}\), and zero at
    \(k_{\text{P}}=k^{*}_{\text{P}}\);
    \item \(g(0) = 0\) for all \(k_{\text{P}}\); and
    \item \(\lim_{b \to -\infty} g(b) = -\infty\) and
    \(\lim_{b \to \infty} g(b) = -k_{\text{P}} < 0\), by
    the same argument as in \autoref{thm:proof_of_kp_star}
    (which holds for all \(k_{\text{P}} \in (0,1)\)).
\end{enumerate}
Since \(b_{\text{crit}}\) is the only critical point, \(g\)
is strictly increasing on
\((-\infty, b_{\text{crit}})\) and strictly decreasing on
\((b_{\text{crit}}, \infty)\). In particular, \(g\) can
cross zero at most once on each of these two intervals.

In the case where \(k_{\text{P}}<k^{*}_{\text{P}}\), we
have \(b_{\text{crit}} > 0\), so the trivial root \(b=0\)
lies in the increasing interval. Since \(g\) is strictly
increasing on \((-\infty, b_{\text{crit}})\) and
\(g(0)=0\), it follows that \(g(b)<0\) for all \(b<0\). On
the decreasing interval \((b_{\text{crit}}, \infty)\), \(g\)
falls from \(g(b_{\text{crit}})>0\) to \(-k_{\text{P}}<0\),
so by the intermediate value theorem and strict monotonicity,
there is exactly one root, which is positive.

In the case where \(k_{\text{P}}>k^{*}_{\text{P}}\), we
have \(b_{\text{crit}} < 0\), so the trivial root \(b=0\)
lies in the decreasing interval. Since \(g\) is strictly
decreasing on \((b_{\text{crit}}, \infty)\) and
\(g(0)=0\), it follows that \(g(b)<0\) for all \(b>0\). On
the increasing interval \((-\infty, b_{\text{crit}})\),
\(g\) rises from \(-\infty\) to \(g(b_{\text{crit}})>0\),
so by the intermediate value theorem and strict monotonicity,
there is exactly one root, which is negative.

Lastly, in the case \(k_{\text{P}}=k^{*}_{\text{P}}\),
\(b_{\text{crit}}=0\) and \(g(b_{\text{crit}})=0\), so only
the trivial root exists, recovering the result of
\autoref{thm:proof_of_kp_star}.
\end{proof}